\pdfoutput=1
\documentclass[acmtog,nonacm,screen]{acmart}

\AtBeginDocument{%
  \providecommand\BibTeX{{%
    \normalfont B\kern-0.5em{\scshape i\kern-0.25em b}\kern-0.8em\TeX}}}

\usepackage{mathrsfs}
\usepackage{graphicx}
\usepackage{xspace}
\usepackage{textcomp}
\usepackage{epstopdf}
\usepackage{wrapfig}
\usepackage{color}
\usepackage{multirow}
\definecolor{turquoise}{cmyk}{0.65,0,0.1,0.1}
\definecolor{purple}{rgb}{0.65,0,0.65}
\definecolor{dark_green}{rgb}{0, 0.5, 0}
\definecolor{orange}{rgb}{0.8, 0.6, 0.2}
\definecolor{red}{rgb}{0.9, 0.0, 0.0}
\definecolor{tomoto}{rgb}{1.0, 0.39,0.28}
\newcommand{\hui}[1]{{\color{black}#1}}

\newcommand{\qian}[1]{{\color{black}#1}}

\newcommand{\etal}{\textit{et al}.\xspace}
\newcommand{\ie}{\textit{i}.\textit{e}.,\xspace}
\newcommand{\eg}{\textit{e}.\textit{g}.\xspace}
\newcommand{\vm}[1]{\mathbf{#1}}

\graphicspath{{./figures-reduced/}}
\begin{document}

\title[Object Properties Inferring from and Transfer for Human Interaction Motions]{\hui{Object Properties Inferring from and Transfer for \\ Human Interaction Motions}}

\author{Qian Zheng}
\affiliation{%
  \institution{Shenzhen University}
}

\author{Weikai Wu}
\affiliation{%
  \institution{Shenzhen University}
}

\author{Hanting Pan}
\affiliation{%
  \institution{Shenzhen University}
}

\author{Niloy Mitra}
\affiliation{%
	\institution{University College London}
}

\author{Daniel Cohen-Or}
\affiliation{%
	\institution{Tel Aviv University}
}

\author{Hui Huang}
\authornote{Corresponding author: hhzhiyan@gmail.com}
\affiliation{%
	\institution{Shenzhen University}
}

\authorsaddresses{%
}



\begin{abstract}
	Humans regularly interact with their surrounding objects. Such interactions often result in strongly correlated motion between humans and the interacting objects. We thus ask:``Is it possible to infer object properties from skeletal motion alone, even without seeing the interacting object itself?''		
	In this paper, we present a fine-grained action recognition method that learns to \textit{infer} such latent object properties from human interaction motion alone. 
	This inference allows us to \textit{disentangle} the motion from the object property and \textit{transfer} object properties to a given motion. 		
	We collected a large number of videos and 3D skeletal motions of the performing actors using an inertial motion capture device. 
	We analyze similar actions and learn subtle differences among them to reveal latent properties of the interacting objects. In particular, we learn to identify the interacting object, by estimating its weight, or its fragility or delicacy.
	Our results clearly demonstrate that the interaction motions and interacting objects are highly correlated and indeed relative object latent properties can be inferred from the 3D skeleton sequences alone, leading to new synthesis possibilities for human interaction motions. Dataset will be available at \url{http://vcc.szu.edu.cn/research/2020/IT}.
\end{abstract}

%
%

		\ccsdesc[500]{Computing methodologies~Computer graphics}
		\ccsdesc[500]{Computing methodologies~Motion processing}

\keywords{motion analysis, human-object interaction, datasets}

\begin{teaserfigure}
	\includegraphics[width=\linewidth]{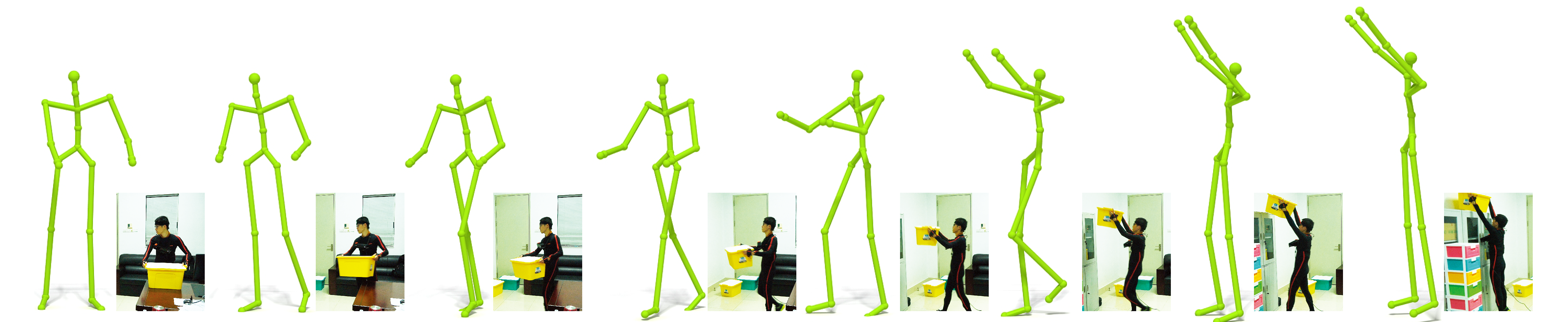}
	\caption{The actor is lifting a box from the table. Can the skeletal motion tell whether the box being lifted is light or heavy?}
	\label{fig:teaser}
\end{teaserfigure}

\maketitle


\section{Introduction}

Digitizing and understanding our physical world are important goals of both computer graphics and computer vision. In natural environments, humans regularly interact with their surrounding objects and, as an effect, such interactions result in strongly correlated motion between humans and the interacting objects. Researchers in experimental psychology show that observers not only can recognize motion categories, but \emph{also infer object properties} by observing corresponding human motion alone, even without directly seeing the object itself~\cite{Blake2007}. For example, we humans, regularly estimate object properties like the weight, fragility, path width, or shape, by observing either the real action of a human or even a pantomimed or virtual avatar action~\cite{Runeson1981,Podda2017,Vaina1995}. 


One way to computationally exploit such correlated human-object motions under interactions would be to learn object properties by learning correlation with human skeletal motion over time. However, the available datasets for human activity recognition~\cite{Shahroudy2016,Liu2017a} are RGB-D videos, which in general contain significant occlusions that hamper the extraction of unseen acting skeletons. While these videos can be used to broadly classify different actions~\cite{LoPresti2016}, we still lack suitable datasets specifically designed for inferring fine-scale variations of object properties.
Unlike previous efforts on action recognition, we analyze \textit{similar actions} and hence have to learn subtle differences among the same type of the action that reveal latent properties of interacting objects. 
Inspired by previous works on motion style transfer, which transform an input motion into a new style while keeping its content, we use these latent properties to edit a given motion.
For example, given the skeletal motion of a person walking on a wide path, we would like to synthesize the person's skeletal motion when walking on a narrow path.

In our work, we focus on \qian{eight} typical types of human-object interaction, including lifting a box, moving a bowl, and walking on a path. 
We collected video and 3D skeletal motions of the performing actors using an inertial motion capture device, which do not suffer from occlusions that are unavoidable from video-based recordings.
For these interactions, we learn to infer latent properties of the interacting object from the 3D skeleton sequences alone.
In particular, we learn to identify the interacting object, by estimating its \emph{property value}, i.e., a particular value of a property, such as 0kg/15kg/25kg for box weight, or empty/full for bowl fragility. 

For the inference task, we treat objects' latent property estimation as a fine-grained classification problem by analyzing similar input skeletal motions. 
\hui{Although some properties (\eg the weight) may vary continuously, treating it as a regression problem requires more training samples.}
We represent a skeleton sequence as a time sequence of graph structure, which encodes the position and speed information of all joints with temporal dynamics. After analyzing per-joint features, we feed it into a recurrent network to recognize the latent object properties. 
The results obtained demonstrate that the interaction motions and interacting objects are highly correlated, where object property values can indeed be inferred, to a certain accuracy, by just observing human movements.
\hui{We will show that, comparing with existing works for action recognition, our method achieves higher inference accuracy.}

For the synthesis task, we develop a network architecture to disentangle  object property from the abstract motion, which allows to create novel skeletal motions by mixing new object properties on target skeletons. \hui{We train a deep neural network with a simple encoder-decoder structure to conduct the disentanglement, \ie the latent space encodes the motion content \textit{without} object property. A motion can then be synthesized given a specific property value.}

In summary, we claim the following contributions:
\begin{itemize}
	\item Learning subtle differences among the same type of motions of humans interacting with an object; 
	\item A property and motion disentanglement network that allows novel motion synthesis conditioned on target interactions;
	\item Introducing an extensive interaction dataset for object property inference from motions with 4k+ samples collected from 100 participants, including eight daily interactions (\ie lifting a box, moving a bowl, walking, fishing, pouring liquid, bending, sitting, and drinking), \hui{which will be released.}
\end{itemize}

\begin{figure*}[t!]
	\begin{center}
		\includegraphics[width=\linewidth]{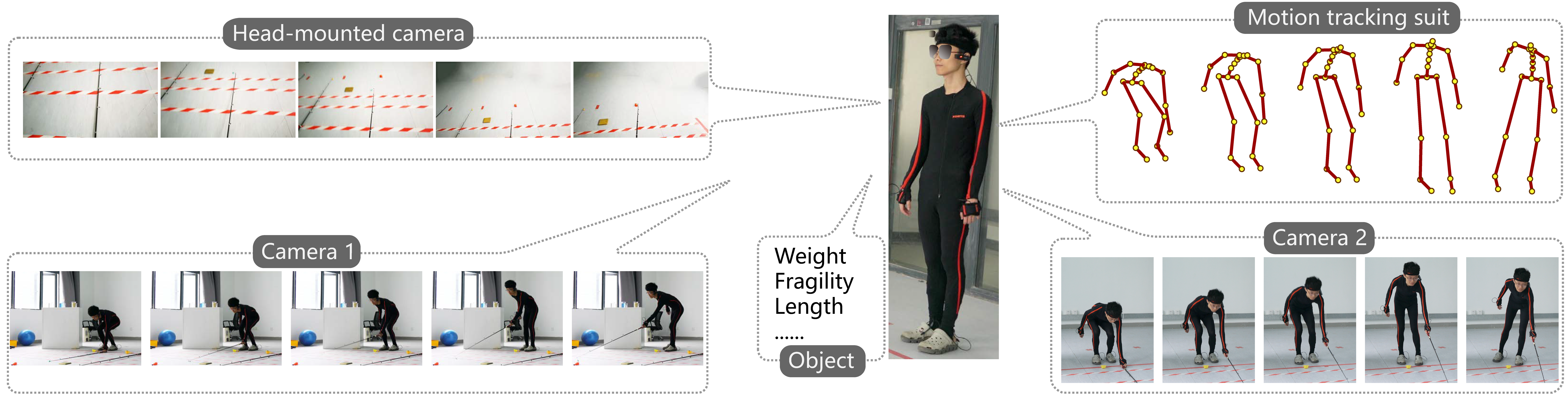}
	\end{center}
	\caption{For each sample, we capture a 3D skeleton sequence by an inertial motion tracking suit, an ego-centric video by a head-mounted camera, two other videos by two cameras placed outside, and the object's geometry along with its properties.}
	\label{fig:capture}
\end{figure*}
\begin{figure*}[t!]
	\begin{center}
		\includegraphics[width=\linewidth]{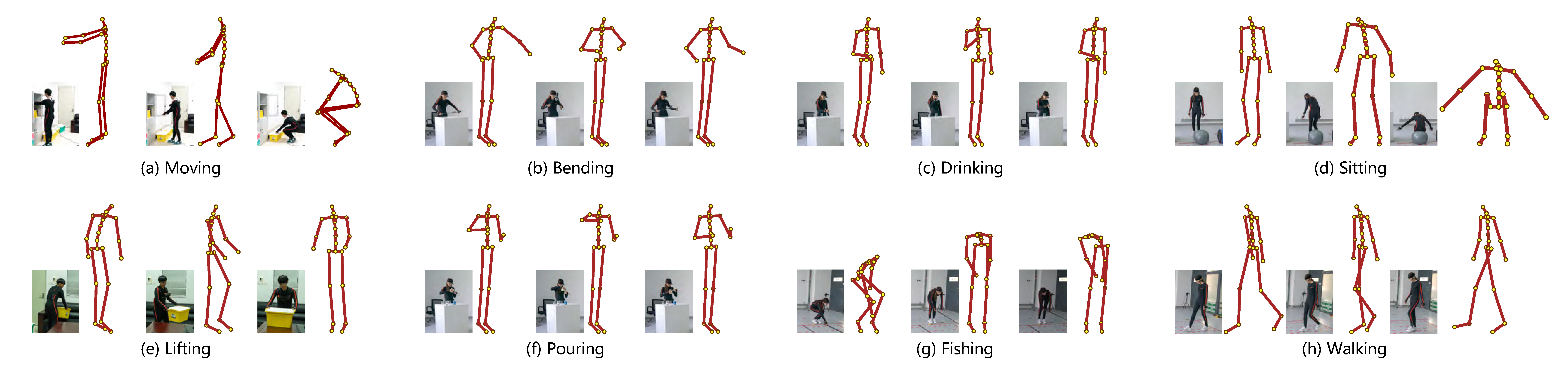}
	\end{center}
	\caption{Eight interaction motions represented in our dataset, which comprises of 4k+ interaction captures across 100 different participants. }
	\label{fig:motion_types}
\end{figure*}

\section{Related Work}

Our work analyzes human interaction motion to detect object properties.
Therefore, we briefly describe previous approaches that exploit human-object interactions from visual inputs, with a focus on object property inference. 
Since we use skeleton sequences to represent motions, we also review those related works on skeleton-based action recognition.

\paragraph*{Human-object interaction.}
Human-object interaction detection itself is an important scientific problem~\cite{Yao2010} with wide practical uses. Recent methods can successfully detect $<$human, verb, object$>$ triplets from visual inputs~\cite{Gkioxari2018,Kato2018}.

A variety of techniques in shape analysis have been developed to extract functional information of objects and scenes using human-object interaction as cues.
An appropriate human pose or action map can be created from an input object~\cite{Grabner2011,Kim2014,Hu2018a} or scene~\cite{Savva2014,Li2019}; see a survey~\cite{Hu2018} for more information.
The hidden human context was used as a cue for labeling and arranging the scenes~\cite{Jiang2013, Jiang2016}.
However, there is no work yet solving this inverse problem: \textit{inferring object properties from human motions and/or interactions alone.}

The spatial relationship between the characters and objects in the environment captures the semantics of interactions.
Ho~\etal~\cite{Ho2010} introduced interaction mesh structure to explicitly represent the spatial relationship for motion retargeting. Later this representation was used for motion comparison~\cite{Shen2019}.

\paragraph*{Object property inference.}
Researchers in psychology reported that observers can make fine discrimination when presented with human motions in visual form. 
The weight of a box can be \emph{seen} by observing another person lifting and carrying it~\cite{Runeson1981}, and the elasticity of a supporting surface can be judged by observing a person walking on that surface~\cite{Stoffregen1994}.
Vaina~\etal~\cite{Vaina1995} demonstrated that the weight of an object was robustly estimated, while size and shape were harder to estimate by observers.
Recently, Podda~\etal~\cite{Podda2017} showed that participants were able to identify the weight of the to-be-grasped object from both occluded real and pantomimed movements, solely using available kinematic information.
Observers seem to focus most on the duration of the lifting movement to perceptually judge the weight~\cite{C.Hamilton2005}. Some findings suggest observers may integrate multiple sources for object property inference; for example, shape, motion, and optical cues are used when inferring stiffness~\cite{Schmidt2017}. Still, we focus on inference from motions alone in this work.

The object classes and their 3D locations can be recovered from motion by exploiting the human-object spatial relations, used for synthetic scene reconstruction~\cite{Kang2017} and scene arrangement recovery~\cite{Monszpart2018}. There is \qian{not much} effort made to automatically infer other properties.  Davis and Gao~\cite{Davis2003} presented a computational framework that can label the effort of an action corresponding to the perceived level of exertion by the performer. Gupta and Davis~\cite{Gupta2007} did a classification of \emph{heavy/light} objects based on the velocity of ballistic motions detected from video.
\hui{Integrating a 3D physics engine is another way to infer physical properties, including mass, position, 3D shape, and friction etc., from real-world videos~\cite{Wu2015,Wu2016}.}

\paragraph*{Action recognition and motion style transfer.}
With the availability of large-scale skeleton datasets, deep learning is popular for action recognition.
Skeleton sequences are indeed the time series of joint positions.
The recurrent neural networks, designed to model long-term temporal dependency problems, have been well exploited for skeleton sequences~\cite{Liu2016,Liu2017,Song2017}. Skeleton is also a special graph structure representation, and thus graph convolution networks are utilized as well for action recognition~\cite{Yan2018}.

CNN models are able to extract high-level information and have also been used to deal with skeleton sequences. A skeleton sequence can be converted into an image or a 3D tensor, and then fed into a CNN to recognize the underlying action. These methods vary most in the representations of skeleton sequences and network structures.
Ke~\etal~\shortcite{Ke2017} represented a skeleton sequence as several images to encode different spatial relationship in-between joints, and then applied pre-trained VGG to extract the features.
Li~\etal~\shortcite{Li2018} represented a skeleton sequence as a 3D tensor, and modeled the global co-occurrence patterns with CNN. 
Most recently, Aristidou~\etal~\cite{Aristidou2018} used a triplet loss network to map short motion clips to an embedding space, where the distances represent similarity between motion clips.
We also utilize graph convolution and RNN to learn object properties from skeletal motions. 
Nonetheless, we propose to use sub-categorical properties to effectively distinguish fine-grained differences between the motions of the same class.

Another related topic is motion style, which usually represents the mood or identity of a particular character's motion. By analyzing differences between performances of the same content in different styles, researchers have proposed the methods to transform an input motion data into new styles~\cite{Hsu2005, Xia2015, Yumer2016}.
The object properties and actions are significantly correlated. A particular object property can be only observable in a particular action type, 
which makes the existing motion style transfer techniques not suitable for our synthesis task.

%

\section{Interaction Motion Dataset Collection}

Traditionally, human motion is captured using optical marker-based systems while the markers are placed on the performer. 
With recent success of deep learning, 2D poses~\cite{Cao2017,Insafutdinov2016,Newell2016,Wei2016,Riza2018} and 3D poses~\cite{Tekin2016,Tome2017,Mehta2017,Kanazawa2018,Pavlakos2018,Andriluka2018} can be extracted directly from RGB or RGB-D video sequences.
Large-scale skeletal motion datasets, such as CMU~\cite{CMU2018}, NTU RGB+D~\cite{Shahroudy2016} and PKU-MMD~\cite{Liu2017a}), are available and driving forces for motion recognition, retrieval and synthesis.
However, although these datasets contain human-object interaction motions, the object information are usually unlabeled, and the (partial) joint trajectories are not sufficient to reliably infer 3D object properties. For example, some limbs are very likely to be occluded by the interacting objects. Such occlusions make it very difficult to robustly extract high-quality skeletal motions from monocular or RGB-D videos, even with state-of-the-art pose detection methods. This is particularly true in our setting where we seek subtle motion differences. 
Therefore,
we use inertial measurement units~(IMUs) to get 3D human motions that are totally free of occlusions.

\paragraph*{Data modalities.} We utilize multiple data modalities to construct our dataset. 
When performing the actions, each subject wore an Xsens MVN inertial motion tracking suit to capture the high-quality 3D skeleton information at 240 frames per second.
Each subject was also required to wear a head-mounted camera to capture ego-centric video. 
Further, we used three uncalibrated cameras to record the subject from three different views, storing three videos at 50 frames per second. 
For each interacting object, in addition to measuring its size and weight, we also scanned its geometry shape.
Figure~\ref{fig:capture} presents our capturing scenario and the data modalities of each motion sample collected.
Although in this work we only use 3D skeletal information to infer the object properties, we believe that these data modalities are useful for the future research.

\paragraph*{Subjects and object interactions.}
We carefully selected human-object interactions to depict the correlation between human motions and properties of objects. For a good candidate, object property values could be inferred easily from the whole interaction motion alone, but difficultly from a single static frame. Following this rule, we chose eight daily interaction: Walking for estimating \emph{the width} of the path, Fishing for \emph{the length} of a fishing rod, Pouring for \emph{the type} of liquid, Bending for \emph{the stiffness} of a power twister, Sitting for estimating \emph{the softness} of a chair being sat on, Drinking for estimating \emph{the amount} of water inside a cup, LiftingBox for \emph{the weight} of an object be lifted, and MovingBowl for \emph{the fragility} of an object. These motions are shown in Figure~\ref{fig:motion_types}. We have invited 100 different subjects for our data collection.
They vary in age (\qian{20--35}), gender (M or F), height (\qian{150--195cm}) and strength (\qian{weak--strong}). 
Here we briefly describe the setting of Walking. Please refer to the appendix for the settings of other interactions.

\noindent 
\textbf{\textsc{Walking.}} Each subject was asked to walk back and forth on three straight paths of different widths. We simulated the width of a path using line markers to indicate path borders, and asked the subjects do not cross the borders. So we have a total of $3\times 2\times 100 = 600 $ motion samples.


\section{Object Property Inference}

\begin{figure}[t]
	\begin{center}
		\includegraphics[width=\linewidth]{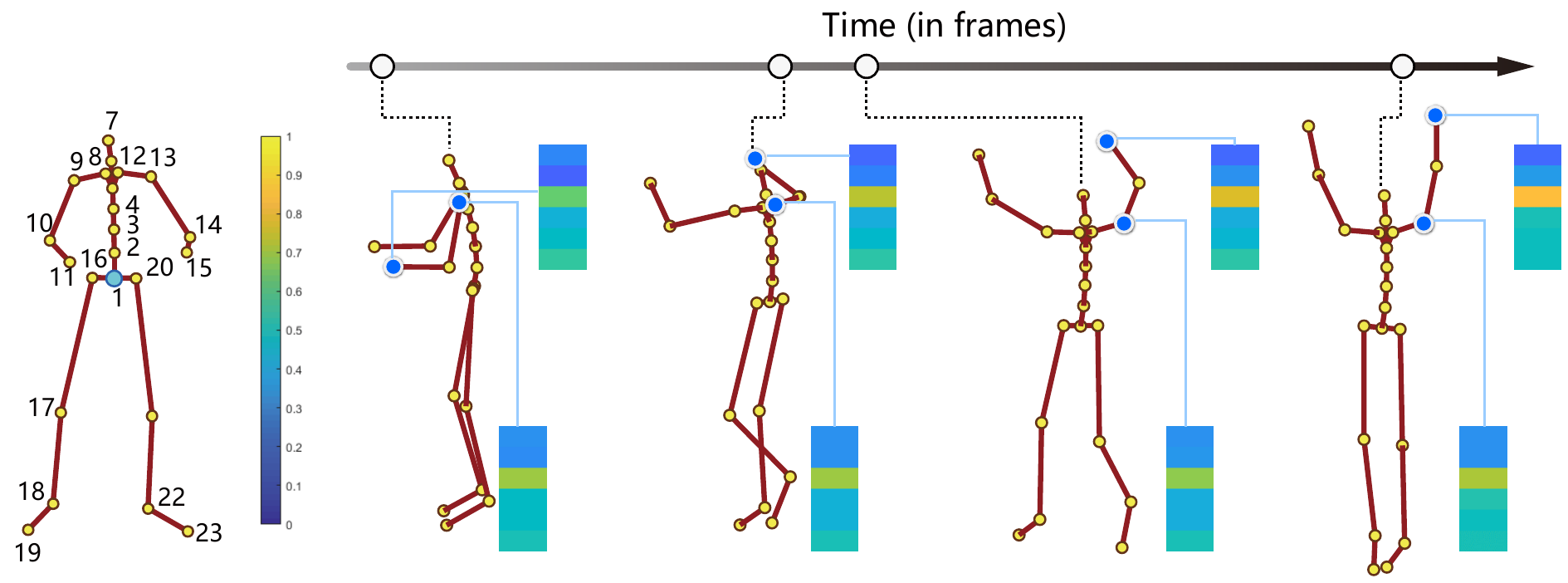}
	\end{center}
	\caption{ We represent a skeleton sequence as a tree sequence. The input feature of each joint is represented by its xyz location and velocity in a local body frame coordinate. The cyan point indicates the root (pelvis) of the tree. Each block indicates the joint's feature at a frame. 
     }
	\label{fig:joint_feature}
\end{figure}

\subsection{Skeleton sequence representation}
The input skeleton data is a sequence of multi-frame tree structure with 3D joints as nodes that form an \textit{action}. As shown in Figure~\ref{fig:joint_feature}, 
a skeleton sequence is denoted as a 3D tensor of size $T\times J\times D$, with $T$  representing the frame length, $J=23$ the total number of joints, and $D$ the feature dimension of each joint, respectively.

Representing a skeleton sequence by joints in xyz locations is common~\cite{Shahroudy2016, Ke2017, Li2018}. Some researchers also represent the joints in 3D angles~\cite{Aristidou2018}. 
\hui{In our case, the object properties that we aim to estimate are highly correlated with the dynamic properties of motions. 
As we show in results, joint trajectories (position and velocity representations) can overall help with object property inference.}

Each joint is represented by the x, y, and z coordinate in a local body coordinate system with its origin on the pelvis joint (indicated with a blue dot in Figure~\ref{fig:joint_feature}).
As local coordinate frame we use, the Z axis to be vertical to the floor, and X axis to be  parallel to the 3D vector from the ``right shoulder'' to the``left shoulder.'' 
For each frame, we use the xyz position relative to the current pelvis joint. Note that in this representation, we ignore the movement of pelvis in the sequence.
We also explicitly encode the velocity of joints. Let the $i$-th joint's position of frame $t$ be $J_i^{t}$. Then, the velocity of a joint $S_i^t$ is approximated as the temporal difference between two consecutive frames: $$S_i^t=(J_i^{t+1}-J_i^t)/\delta{t},$$
while $\delta{t}$ represents the time interval between consecutive frames.



\begin{figure}[tb!]
	\begin{center}
		\includegraphics[width=\linewidth]{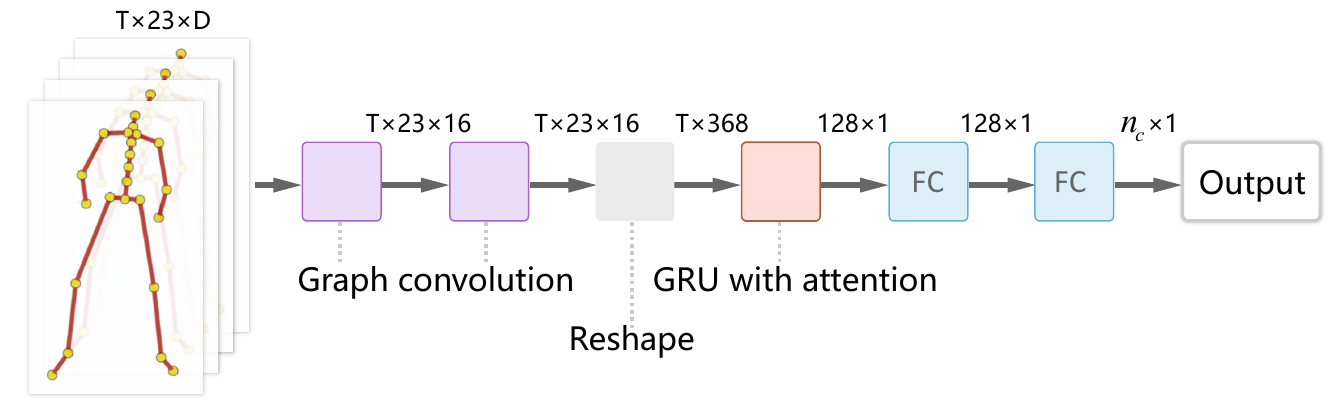}
	\end{center}
	\caption{We represent a skeleton sequence by a 3D tensor of size $T\times J\times D$, $T$ representing the frame length, $J$ the number of joints, and $D$ the feature dimension of each joint, respectively. Our classifier for object property values is made of graph convolution layers, GRU, and fully connected layers. The size of \qian{the tensor} after each layer is indicated in the figure with $n_C$ denoting the number of classes for an object property,
		\eg, $n_C$=6 when the input is a lifting motion and the object property is the weight of box being lifted.}
	\label{fig:network}
\end{figure}

\subsection{Object property classifier}
In practice, our object property classifier consists of two graph convolution layers, a GRU layer~\cite{cho2014learning}, and then two fully connected (FC) layers for the final classification, i.e., the object property inference;
see Figure~\ref{fig:network}. 
The graph convolution layer computes the per-joint features considering the known human body skeleton topology. 
The GRU layer with attention accumulates the information of all frames and computes the importance of each joint. 
The combination of graph convolution layers and GRU units enables us to better infer object property values from the same types of motions.

\paragraph*{Graph convolution layer.}
Graph convolution usually deals with the undirected graph. As the skeleton is a hierarchical tree structure, for a given joint, we only consider its parent, instead of all neighbors, to apply a convolution. 
Formally, for the $i$-th joint of frame $t$, its feature after graph convolution $\vm{x}_{t,i}'$ is:
\begin{equation}
\vm{x}_{t,i}' = \vm{Relu}
\left(\vm{W}_g 
\begin{bmatrix}
\vm{x}_{t,i}  \\
\vm{x}_{t,j}-\vm{x}_{t,i}
\end{bmatrix} + \vm{b}_g
\right), 
\end{equation}
where $\vm{x}_{t,i}$ represents the feature of this joint fed to this layer, $j$ is its parent's index, and $\vm{W}_g,\vm{b}_g$ are the learnable weights for a graph convolution layer.
Experiments clearly show that using skeleton topology information can
improve the inference accuracy; see e.g., Figure~\ref{fig:graph_conv}. We use this asymmetric edge function as suggested in \cite{Wang2019}.  


\paragraph*{The GRU layer with attention.}
Attention mechanics is widely used in skeleton-based action recognition. It can improve action recognition and discover the relative importance of joints and frames.  
For example, Zhang~\etal~\shortcite{zhang2018} use an element-wise attention gate to a RNN block to improve action recognition. We also add a joint-wise gate to the RNN cell. The attention value of each joint of frame $t$ is computed based on the hidden state of the RNN cell $\vm{H}_{t-1}$:
\begin{equation}
a_{t,i} = \vm{sigmoid}(\vm{W}_{h}\vm{H}_{t-1}+\vm{W}_{x}\vm{x}_{t,i}+\vm{b}_a), 
\end{equation}
where $\vm{x}_{t,i}$ represents the feature of the $i$-th joint fed to the RNN cell, and  $\vm{W}_{h}, \vm{W}_{x}, \vm{b}_a$ are the learnable weights for an attention convolution layer.
Then, the input fed to the RNN cell is updated as $\tilde{\vm{x}}_{t,i} = (1+a_{t,i})\vm{x}_{t,i}$, where $a_{t,i}$ represents the importance of $i$-th joint at frame $t$.

\paragraph*{Implementation details.}
For all experiments presented here, we use $J=23$ major body joints.
We use the classic cross entropy loss as it is a classification problem.
For skeletal representation, we apply a normalization
pre-processing step. 
%
The lengths of collected motion samples vary from 3s to 6s. 
Additionally, we used data augmentation to increase the
number of samples \hui{and to remove the rotation bias}. 
We rotated each sequence along the Z axis 10 times and cropped 10 sub-sequences from each original and rotated sequence. 
The rotation angles were drawn from a uniform distribution between $[0, \pi)$, and the cropping ratios were drawn from a uniform
distribution $U[0.9, 1]$.
This data augmentation enlarged the size of our skeletal motion dataset by 100 times.
We down-sample each sub-sequence to 30 frames.
We used TensorFlow with the network initialized with Adam optimizer with a batch size of 32 and a learning rate of 0.0001.
Training was stopped after 60 epochs by default.

\begin{figure}[t]
	\begin{center}
		\includegraphics[width=\linewidth]{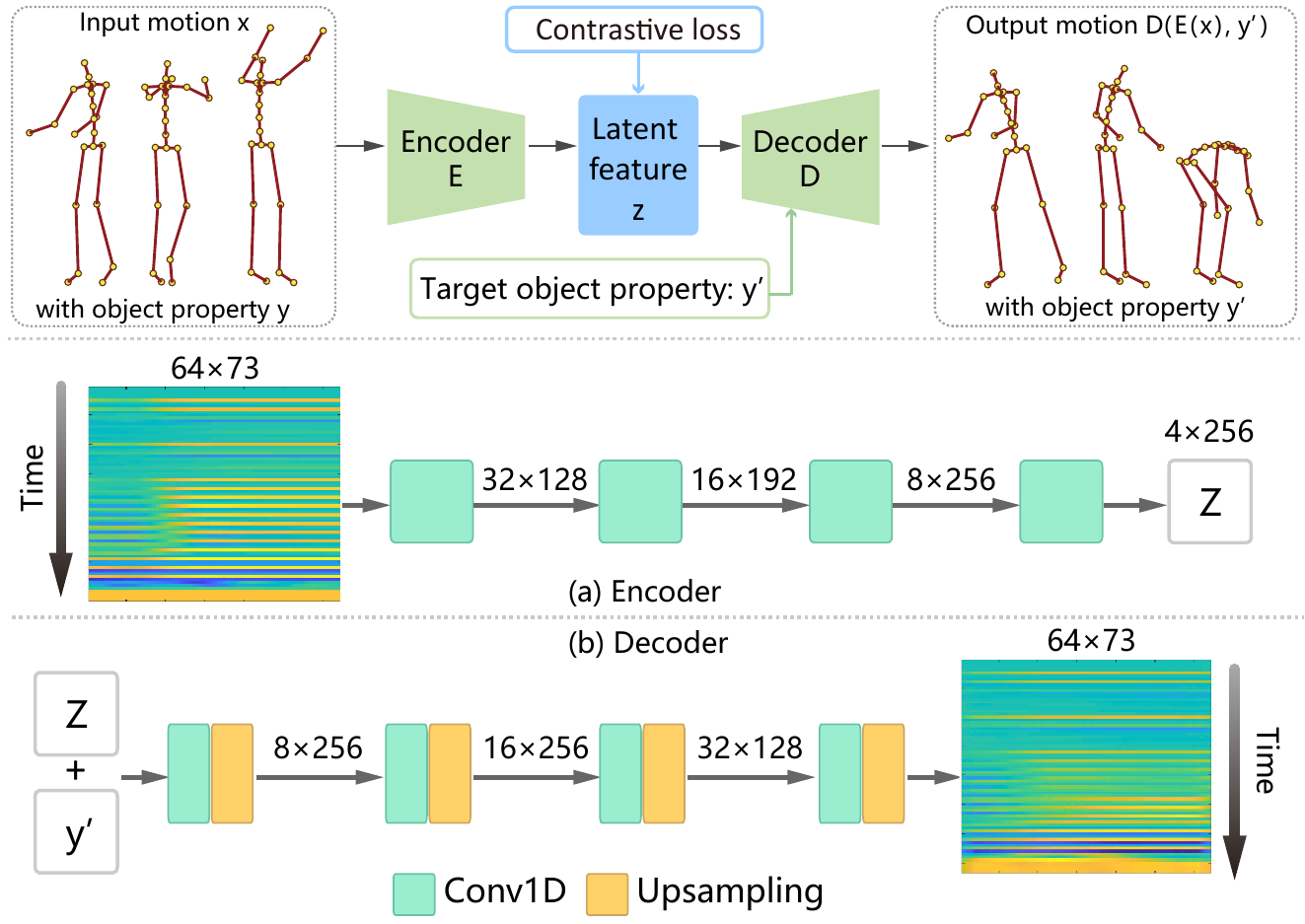}
	\end{center}
	\caption{\hui{Network for motion transfer driven by object properties. That is, by changing the object property value $y$, we may generate human motions that match well with the given property value.}}
	\label{fig:transfer_network}
\end{figure}

\begin{figure}[t]
	\begin{center}
		\includegraphics[width=\linewidth]{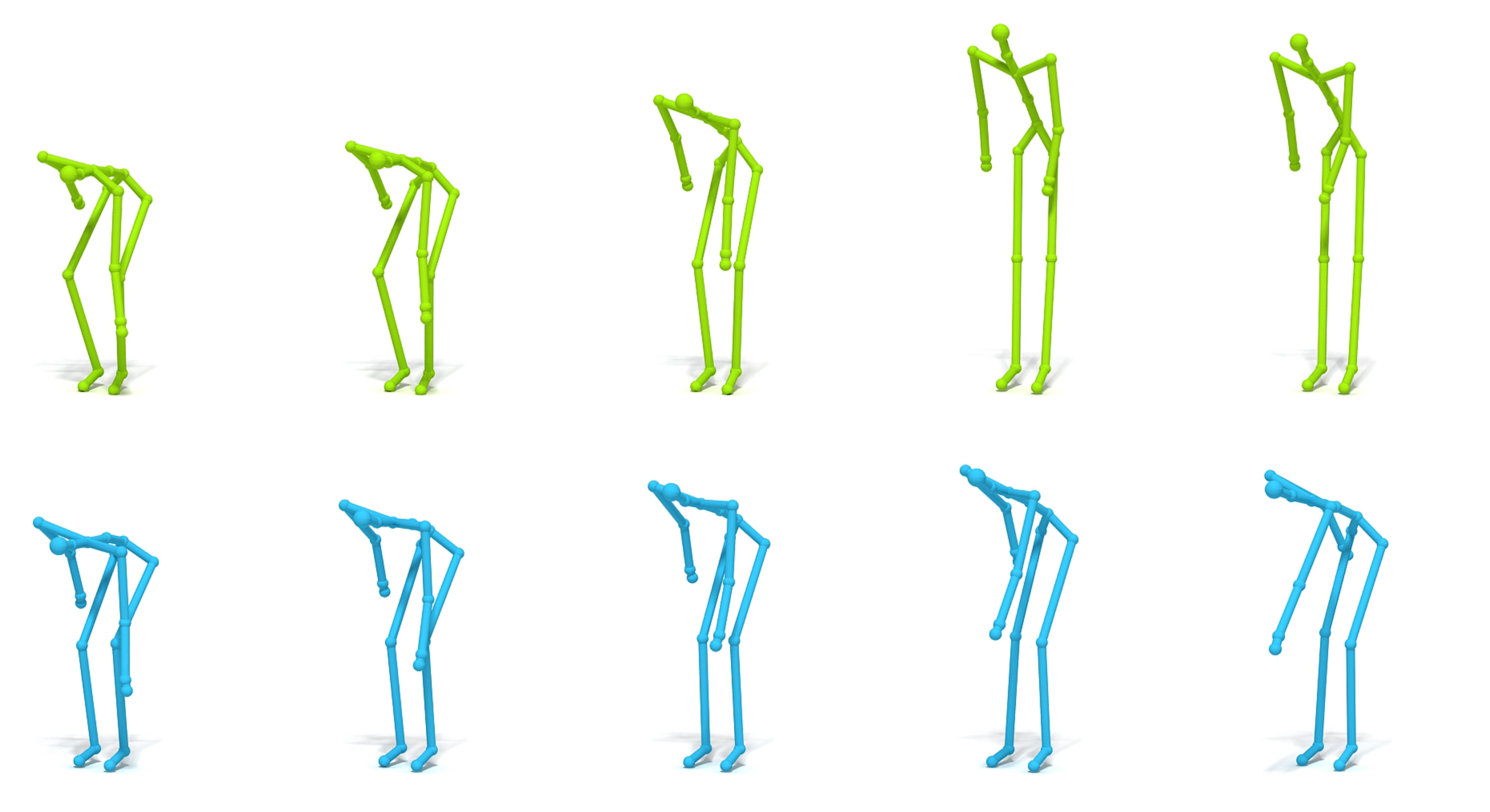}
	\end{center}
	\caption{Given a fishing motion with a long rod (the green), we transfer the rod from \emph{long} to \emph{short} to get a new motion (the blue).} 
	\label{fig:fishing_transfer}
\end{figure}

\section{Object Property-aware Motion Transfer}
In the synthesis content, our goal is to use target object property values to guide motion transfer for a given actor. Given an interaction skeletal motion $x$ whose object property value is $y$, and a new target object property value $y'$, we want to generate new skeletal motion $x'$ that matches the given target property value $y'$.

Inspired by~\cite{Holden2016,Aberman2019}, we use an encoder-decoder structure to perform this motion retargeting; see Figure~\ref{fig:transfer_network}. The encoder $E$ converts an input motion to a latent space $z=E(x)$, and the decoder $D$ synthesizes a new motion conditioned on the target property value, denoted as $D(E(x), y')$.
To train the network, we use a loss function consisting of two terms: a reconstruction loss and a contrastive loss.

The {\em reconstruction loss} aims to constrain the encoder and decoder. We want the output motion to be similar to the motion performed by the same subject under the target property value $y'$, denoted by $\hat{x}$. When $y'$ equals $y$, $\hat{x}$ equals $x$. We use the Euclidean loss in the local coordinate frame to measure the quality of the reconstruction:
\begin{equation}
\mathcal{L}_{rec} (E,D)=\mathbb{E}_{x,y'}\|D(E(x),y')-\hat{x}\|^2_2. 
\end{equation}
The exact choice of the reconstruction loss is not fundamental here. Other reconstruction loss especially designed for motion
frames, such as geodesic loss measuring the 3D rotation errors of joints~\cite{Gui2018}, could be used.

Another loss is a {\em contrastive loss} that ensures that $E(x)$ does not have residual information about the input object property~\cite{Hadsell2006}:
\begin{equation}
\begin{split}
\mathcal{L}_{ctr}  (E)= \mathbb{E}_{x, x^+} {\|E(x)-E(x^+)\|^2_2} +
\mathbb{E}_{x, x^-}{\big[\alpha-\|E(x)-E(x^-)\|_2 \big]_{+}^2}.
\end{split}
\end{equation}
To help disentanglement, we constrain the distance in latent space between different motion samples. 
Taking an anchor motion $x$, we compare it with a positive motion $x^+$ that comes from the same performer under a {\em different} object property value, and a negative motion $x^-$ that coming from a different performer under the {\em same} property value.
The dissimilarity between the anchor motion and negative motion should be larger than a margin $\alpha$, and the distance between the anchor motion and positive motion should be small. The full objective functions to optimize the encoder $E$ and decoder $D$ is a combination of two terms:
\begin{equation}
\mathcal{L}(E,D) = \mathcal{L}_{rec}(E,D)  + \lambda \mathcal{L}_{ctr}(E), 
\end{equation}
where $\lambda$ is a hyper-parameter that controls the relative importance of contrastive loss compared with the reconstruction loss. We use $\lambda=0.1, \alpha=5$ in all our experiments.

Here the skeleton sequence for motion transfer is represented by the local and global motion as suggested in~\cite{Holden2016},  which is slightly different from that for object property inference.
For local motion, we use joints in XYZ locations of a local frame coordinate,
just as the representation for property inference.
Global motion consists of the root's global velocity and foot contact labels.
See Figure~\ref{fig:transfer_network}; the rows represent the location of a joint over time.
We down-sample the motion to 64 frames.

The encoder is composed of 4 1D convolutional layers with the stride size of two for down-sampling the time axis.
The decoder is composed of 4 nearest-neighborhood up-sampling followed by convolution of stride 1 to restore the motion; see Figure~\ref{fig:transfer_network}. 

All models are trained using Adam with $\beta_1=0.9$ and $\beta_2=0.999$. The batch size is set to 32 for all experiments.
We train all models with a learning rate of 0.00001.
Training takes about 10 minutes on a server with an Intel Xeon
2.20GHz CPU 10 cores, 256GB memory, and a NVIDIA TitanXP GPU.

\begin{table}[b!]
	\caption{Object property inference accuracy (\%) on the cross-subject settings. The weight has 6 classes (0, 5, 10, 15, 20, and 25kg). The fragility has 3 levels, implicitly reflected by moving without spill-over an empty bowl, a bowl full of rice and a bowl full of water, respectively. The width of the path, length of the rod, type of the liquid, stiffness of the power twister and water amount in the cup also have 3 levels. The softness of the chair has 4 classes.}  
	\label{tab:accuracy1}
	\resizebox{\linewidth}{!}{
		\begin{tabular}{r|cc}
			\toprule
			\multirow{2}{*}{Object property} 	
			& \multicolumn{2}{c}{Accuracy (\%)}  \\
			 & Ours & ST-GCN \\
			\midrule
			Lifting a box for weight (6) & \textbf{61.8} & 57.3 \\ 
			Moving a bowl for fragility (3) &  77.5 & \textbf{78.9} \\
			Walking for path width  (3) & \textbf{83.9} & 73.8\\
			Fishing for length of rod (3) & \textbf{80.7} & 77.2\\
			Pouring for type of liquid (3) & \textbf{62.8} & 62.1\\
			Bending for stiffness (3) & \textbf{71.6} & 44.7\\
			Sitting for softness of chair (4)& \textbf{73.7} & 66.4\\
			Drinking for water amount inside the cup (3) &  \textbf{62.5} & 57.0\\
			\bottomrule
		\end{tabular}}	
\end{table}

\begin{figure*}[t!]
	\begin{center}
		\includegraphics[width=\linewidth]{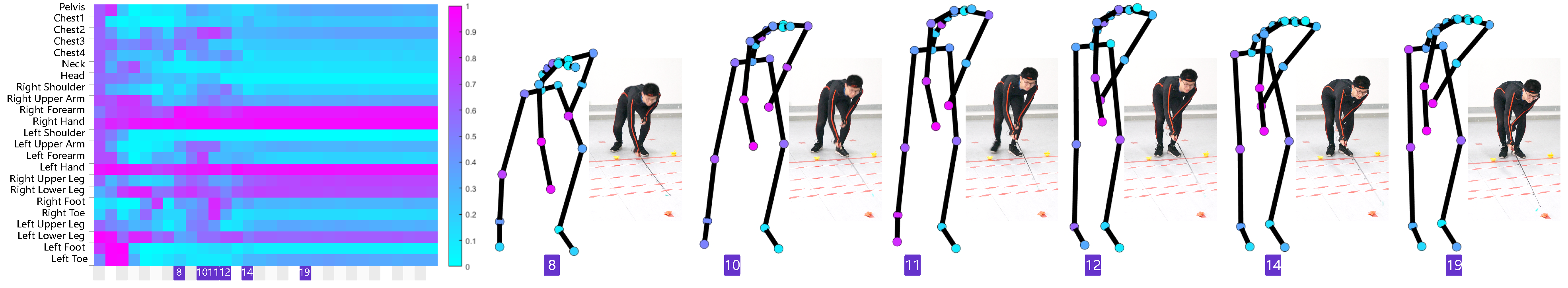}
	\end{center}
	\caption{Estimating the joint-level importance of a fishing motion for inferring the object property. Note that here the color of magenta to cyan indicates the importance from high to low.} 
	\label{fig:joint_attention}
\end{figure*}

\begin{figure*}[t]
	\begin{center}
		\includegraphics[width=\linewidth]{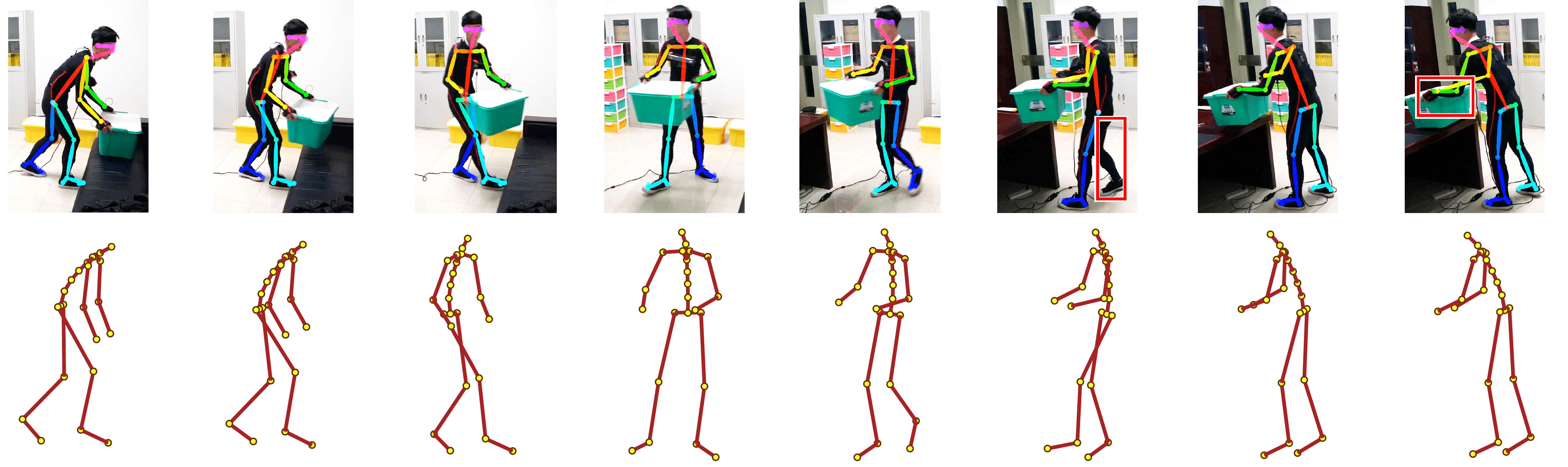}
	\end{center}
	\caption{We show some 2D skeletons extracted from our recorded video at the top, where missing parts are highlighted with red boxes. In comparison,  3D IMU skeletons captured at the corresponding frames are shown underneath, which are clean and complete.}
	\label{fig:2d_skel}
\end{figure*}

\section{Results and Evaluation}
\subsection{Evaluation for object property inference}

To measure the model performance on the object property inference, we conducted a cross-subject evaluation.
We split the 100 participants into training (60), validation (20), and testing (20) groups, respectively. 
\hui{Hence the testing is done with different people rather than the ones who were employed for training and validation.}
During training, we select the network parameters with the smallest validation error among all the iterations. Then, we evaluate and report performance on the testing groups.

\hui{
We implemented several variants to evaluate the impact of different skeleton representations. As using both position and speed achieves the best performance, we applied this representation on other tests.
We reported the object property inference accuracy on all eight types of motions. To evaluate, we used a state-of-the-art method for action recognition based on skeletons to set a baseline.
We also evaluate the utility of the graph convolution layer and GRU units with attention.
Furthermore, we test the inference
accuracy regarding the sensitivity of the object property difference.
}

Table~\ref{tab:accuracy1} shows the object property inference accuracy (\%) on the cross-subject settings.
\hui{The performance looks not very impressive by a first glance at the numbers. Nonetheless, in consideration of the subtle difference among motions under different object properties, we believe this accuracy is reasonable. Furthermore, in most cases, our method outperforms the baseline. We describe the detail of lifting motion in the following as an example.}
Lifting motion is for the weight estimation from human interaction motions. We trained a classifier that outputs 6 classes corresponding to the weights from 0kg to 25kg with a step of 5kg.
The accuracy is about 62\% on the cross-subject setting. 
Considering that the weight difference among the classes are relatively small and the lifting motion is also highly related to the strength of the performer, the resulting estimation accuracy is effective for such subtle changes.

\begin{table}[tb]
	\caption{Impact of different skeleton representations for the inference accuracy (\%) on the cross-subject setting.} 	
	\resizebox{\linewidth}{!}{
		\begin{tabular}{r|ccc}
			\toprule
			& Lifting (6) & Walking (3) & Fishing (3)\\ 
			\midrule
			Position & 57.82   & 76.84 & \textbf{84.21} \\  
			Euler angles & 43.38  & 81.58  & 73.68  \\
			Speed  & 59.93  & 79.82  & 69.4 \\
			Angular speed & 47.46 & 73.16 & 63.51 \\
			\midrule
			Position, Euler angles &  55.70  & 79.65  & 71.58 \\
			Position, speed & \textbf{61.81} & \textbf{83.93} & \textbf{80.70} \\			
			Position, angular speed  & \textbf{64.58} & 79.47 & 77.54 \\
			Speed, angular speed & 55.70  & \textbf{84.39}  & 76.49  \\
			Speed, Euler angles, & 50.56 & 70.00 & 66.67 \\
			Euler angles, angular speed & 56.06 & 80.53 & 72.28 \\
			\midrule
			Position, Euler angles, angular speed & 50.35 & 78.42 & 70.18 \\
			Position, speed, angular speed & \textbf{62.32} & \textbf{82.98} & \textbf{78.95} \\
			Position, Euler angles, speed &  56.55 & 81.58 & 71.93 \\
			\midrule
			Position, Euler angles, speed, angular speed & 58.73 & 82.98 & 78.95 \\ 
			\bottomrule			
		\end{tabular}}
	\label{tab:eval_skl}
\end{table}

\begin{figure}[t!]
	\begin{center}
		\includegraphics[width=\linewidth]{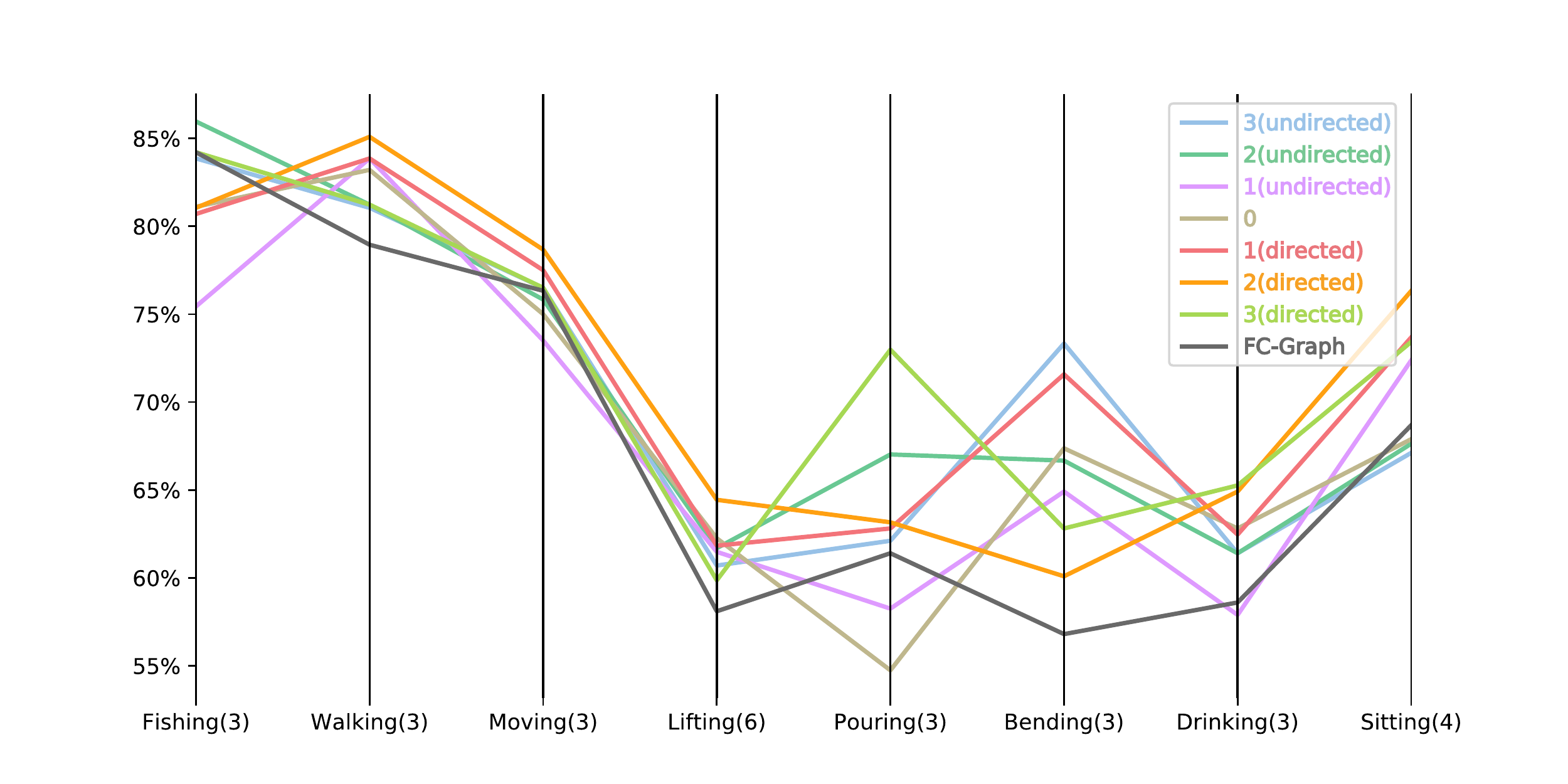}
	\end{center}
	\caption{Parallel coordinates representation for inference accuracy with different ways of computing per-joint features in the 2 graph convolution layers. Each vertical axis represents the inference accuracy from a type of motion. Each line represents a setting. Considering all motions types, it seem good to use the parent of a joint to compute joint feature (the red line).}
	\label{fig:graph_conv}
\end{figure}

\paragraph*{Baseline.}
We used a state-of-the-art method for action recognition based on skeletons~\cite{Yan2018} (denoted by ST-GCN) to be a baseline to evaluate the fine-grained motion inference.
ST-GCN consists of 9 layers and has about 0.3 million parameters, which is about ten times larger than our model. 
\hui{The original network performed very poorly probably due to the small size of our motion dataset. Setting the layer number as three achieved the best performance during our tuning. We thus reduced the original ST-GCN to three layers. This also leads to a similar parameter setting as ours.}
%
%
We also used both position and speed to represent the skeletal motion. The last column in Table~\ref{tab:accuracy1} shows its performance on the cross-subject setting. Overall speaking, our proposed method has achieved higher inference accuracy.

\paragraph*{Choices of skeleton representation.}
To evaluate the impact of skeleton representations, we tried several variants. A skeleton sequence was represented by the positions of joints, or the rotation matrix of bones. Similarly, the motion dynamic was measured by the joint speeds or bone angular speeds.
We represented the skeleton sequence by different forms, and then evaluated their performance on object property inference of three different motions (i.e., lifting, walking, fishing). All other settings were exactly the same. 
Table~\ref{tab:eval_skl} shows that the best representation varies for different object properties. Yet overall speaking, using both position and speed is a good option. So this representation was used in other experiments.

\paragraph*{Graph convolution.} 
To evaluate the impact of the graph convolution layer regarding per joint feature, we fixed other layers and only changed the two graph convolution layers, and report its performance on object property inference; see Figure~\ref{fig:graph_conv}. We evaluated on different settings: ignoring the connections between joints and only considering the joint itself to compute per joint feature (similar to PointNet~\cite{Charles2017}), or treating the skeleton as a tree whose root is the pelvis (directed graph), or treating it as an undirected graph. We also considered different numbers of ancestors (from $1$ to $3$) of each joint. For an undirected graph, we also considered its k-degree neighborhoods using $k=1,2,3$, or all nodes (FC-Graph) in our tests. 
Figure~\ref{fig:graph_conv} shows that though the inference performance varies across the types of motions, considering a joint's parent to compute its feature is a good option.

\paragraph*{Joint-level attention.}
The learned attentions marginally improved the object property inference, especially for the rod length inference from the Fishing and the softness of chair inference from Sitting motion, both increased about 4\%.
We visualized the attention weights on joints by the color. For better visualization, we linearly mapped the squared attention values to colors to highlight the importance. Figure~\ref{fig:joint_attention} shows the attention weights on the two arms are large for the \emph{fishing} motion, consistent with our human intuition.

\paragraph*{Weight and water amount sensitivity.}
To evaluate the inference accuracy regarding to the sensitivity of the object property difference, we trained and tested the model with several different subsets of motion samples, i.e., using samples with only some specific property values. For example, when evaluating the model's ability to distinguish 5 kg from 10 kg, only motion samples with these two weights were used. All other settings were exactly the same.

\begin{table}[tb]	
	\caption{The object weight and water amount inference accuracy (\%) under different configurations: two, three, or six classes. See text for more details. } 
	\begin{center}
		\begin{tabular}{cccc|cc}
			\toprule
			\multicolumn{4}{c|}{Lifting~(kg)} 
			& \multicolumn{2}{c}{Drinking} \\
			
			5/25 (2) & 10/15 (2) & 5/15/25 (3) & (6) & Empty/Full (2) & (3) \\
			\midrule
			94.7 & 78.7 & 81.7 & 61.8 & 86.8 & 62.5 \\
			\bottomrule	
		\end{tabular}		
	\end{center}
	\label{tab:eval_weight}
\end{table}

Table~\ref{tab:eval_weight} shows that the inference performance is related with the weight label distribution. Note that 2-class classification accuracy drops dramatically from 94.7 down to 78.7 when classifying 10/15kg boxes instead of 5/25kg, even lower than the 3-class classification accuracy of classifying 5/15/25kg. 
We argue that this is mainly caused by the small dynamic motion difference when lifting boxes are close in weight.
The water amount label distribution also shows a similar trend.


\subsection{Comparison with videos}
\paragraph*{Property inference from only videos.}
We additionally evaluate the weight and fragility inference performance from different input sources.
In particular, we have tested the performance using 2D skeleton sequences directly extracted from videos that were recorded from a fixed view. 
We used OpenPose detector~\cite{Cao2017} to extract 25 body keypoints in 2D to get image-space skeletons using videos.
Due to the fixed camera view and the occlusion of interacting objects, extracted 2D skeletons may have large missing parts in some frames; see e.g., Figure~\ref{fig:2d_skel} (top). 
We choose the most representative 17 body joints, and replace the 3D IMU skeletons with corresponding 2D video skeletons. Now the skeleton sequences have only x and y positions without z dimension. 
The speed and acceleration attributes are not used as there are unavoidable flickers in video sequences and they cannot be easily lifted to 3D.

\begin{figure}[tb!]
	\begin{center}
		\includegraphics[width=0.49\linewidth]{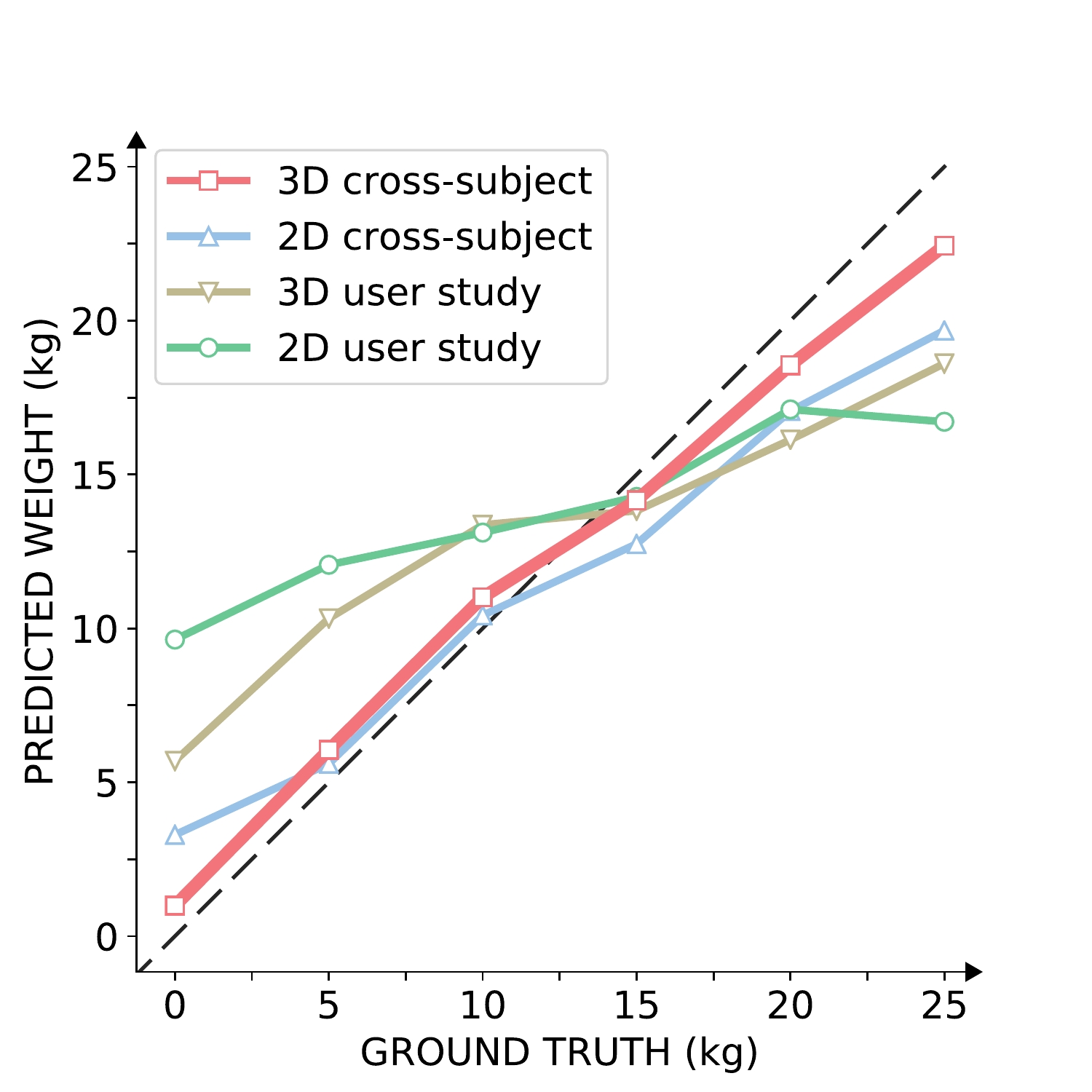}
		\includegraphics[width=0.49\linewidth]{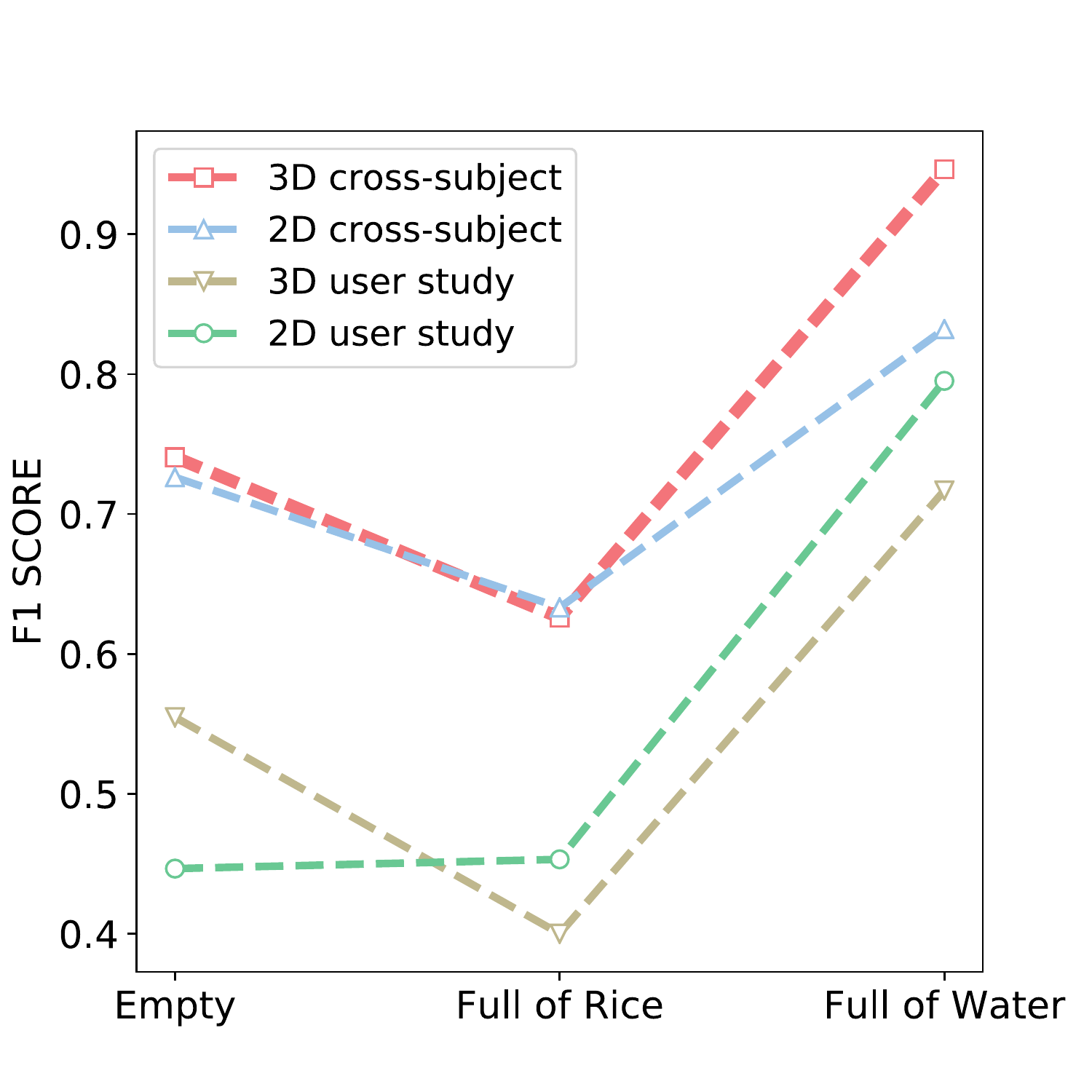}
	\end{center}
	\caption{Left: The average of predicted weights by our model and human observers on both 2D and 3D skeletal cases, where the weights vary among 0, 5, 10, 15, 20, and 25kg. The 6-class weight predictions by our model using 3D skeletons are much closer to ground truth indicated by the slant black line. Right: The F1 score per class of the fragility estimation by our model and human observers. On both 2D and 3D skeletal cases, our method (see the red and blue marks) achieves better results. }
	\label{fig:weight_fragility}
\end{figure}

Figure~\ref{fig:weight_fragility} presents the evaluation of 6-class weight classification and 3-class fragility inference on cross-subject settings, by our model trained on 2D and 3D skeletons and human observers.
Using 2D skeletons instead of 3D causes some drop in inference accuracy in both weight and fragility estimation, see the red and blue lines. 
We believe this is mainly due to joint estimation errors, depth information missing, and kinematic flicker artifacts.


\begin{table}[t]
	\caption{\hui{When adding rendered skeletons into training, the object inference accuracy (\%)  (such as the weight by lifting and the fragility by moving) from videos can be improved significantly as compared below.}} 
	\label{tab:eval_rendered_skls}
	\begin{tabular}{lcc}
		\toprule
		& Lifting (6) & Moving (3)\\
		\midrule
		without &  51.6 &  62.9  \\
		with &  \textbf{61.4} & \textbf{71.4} \\
		\bottomrule
	\end{tabular}
\end{table}

\begin{figure*}[t!]
	\begin{center}
		\includegraphics[width=\linewidth]{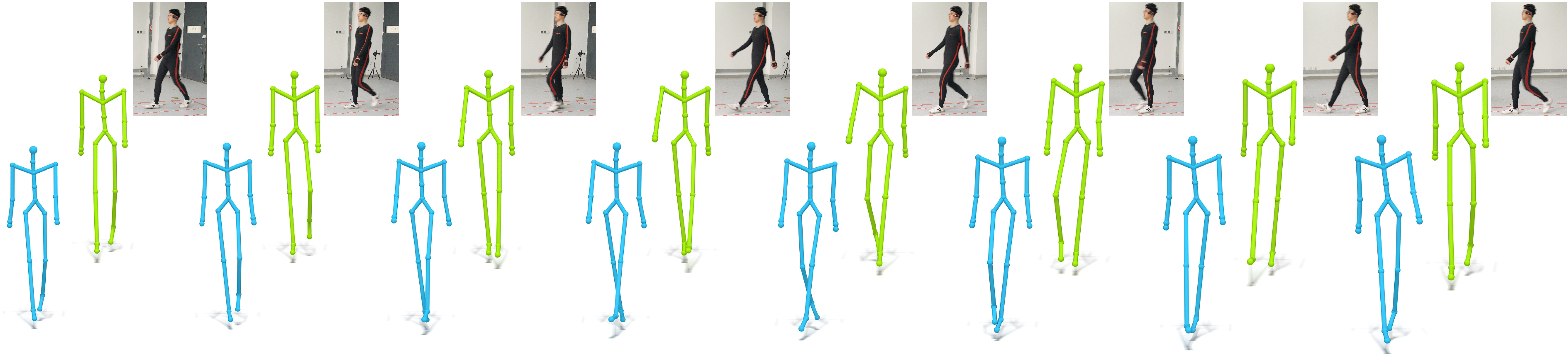}
	\end{center}
	\caption{Given a motion sequence of an unseen subject walking on the wide path (in green), we can generate a new sequence that looks like the subject was walking on a narrow path (in blue).}
	\label{fig:walking_transfer}
\end{figure*}

\begin{figure*}[t!]
	\begin{center}
		\includegraphics[width=\linewidth]{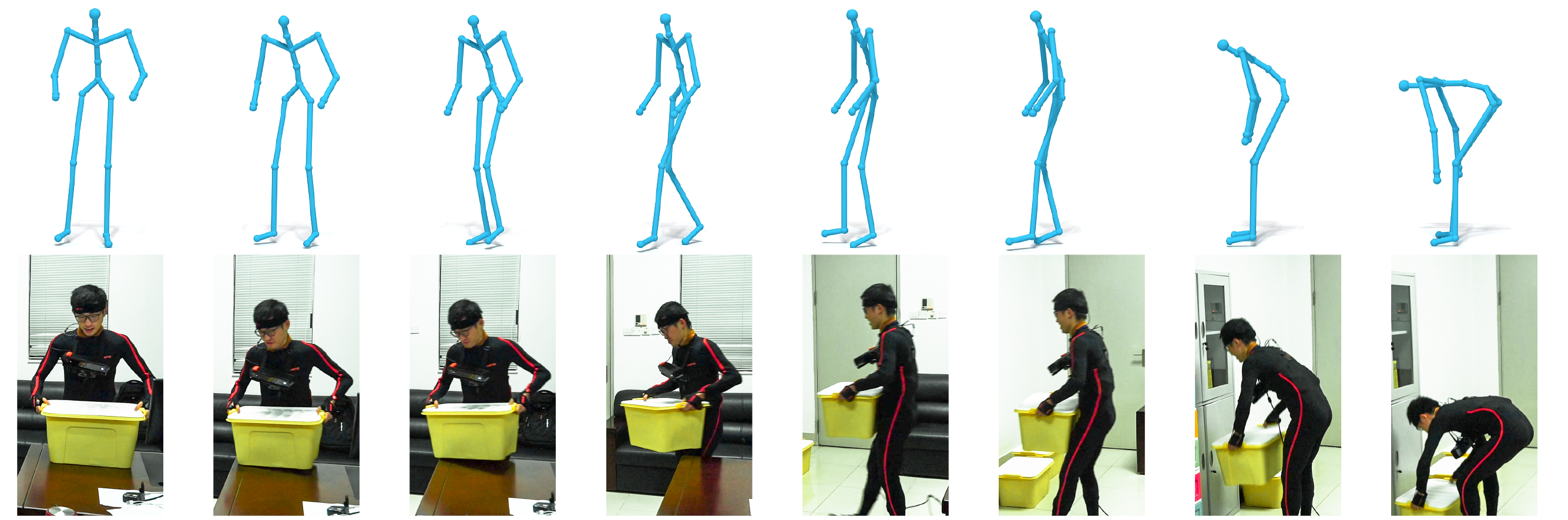}
	\end{center}
	\caption{\hui{Given the motion sequence shown in Figure~\ref{fig:teaser}, we can generate a new sequence that looks like the subject was lifting a heavy box, but it was too heavy to be lifted. The generated motion is similar to the ground truth as shown with a sequence of RGB images at the bottom.}}
	\label{fig:lifting_transfer}
\end{figure*}

\begin{figure}[tb!]
	\begin{center}
		\includegraphics[width=\linewidth]{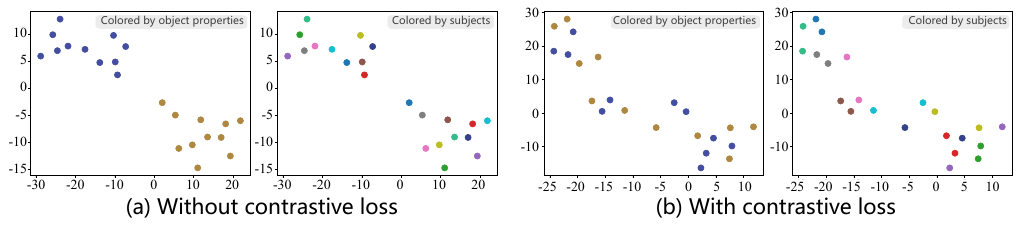}
	\end{center}
	\caption{Latent variables after encoder of several lifting motions with 0 kg and 25 kg boxes are projected to 2D space. Without contrastive loss (a), the left is colored by object properties, and the right by subjects. With contrastive loss (b), colored the same way.}
	\label{fig:latent_x}
\end{figure}

\paragraph*{Property inference from videos enhanced by 3D skeletons.}
The small size of unoccluded 3D skeletons motion samples may generate \qian{thousands} of rendered 2D skeletons. Here we show these 2D projections of 3D data can effectively improve the performance of property value estimation from 2D videos.
We generated these virtual 2D samples by projecting the 3D joint positions of 3D skeleton sequences according to different camera view angles. For the virtual camera setting, we used a weak-perspective camera model, as suggested by~\cite{Aberman2019}, which generates 2D projections of synthetic 3D skeleton sequences. For every 3D sequence, we used 8 fixed views, placed a camera every 22.5 degrees around the actor (covered about 180 degrees in total), and all cameras were set to be horizontal (pitch angle equals to $0$).

Table~\ref{tab:eval_rendered_skls} presents the evaluation of 6-class weight classification and 3-class fragility inference on the cross-subject setting, by our models trained on 2D skeletons extracted from videos only, or on 2D extracted skeletons and rendered 3D skeletons. The trained models were tested only on 2D extracted skeletons. In the second case, The ratio of extracted and rendered skeletons was $1:8$. Clearly using additional virtual skeletons can effectively improve the performance.

\subsection{Evaluation for property-aware motion transfer}
We again split the 100 subjects into training (60), validation (20), and test (20) groups, respectively.
During training, we select the network parameters with the smallest validation error among all the iterations. We evaluate and report performance on the test groups.

\paragraph*{Latent space visualization.}
Figure~\ref{fig:latent_x} shows the latent space of motion samples after projecting the latent features to a 2D image using t-SNE. Each point represents a motion sample of a subject lifting a 0 kg or 25 kg box. 
The leftmost figure shows that they are clustered according to object property values without contrastive loss.
This is due to the motion differences among different subjects are smaller than that of lifting 0 kg and 25 kg boxes.
With the contrastive loss, the features start to disentangle from object properties and become more related to the subjects.

\begin{figure*}[t!]
	\begin{center}
		\includegraphics[width=\linewidth]{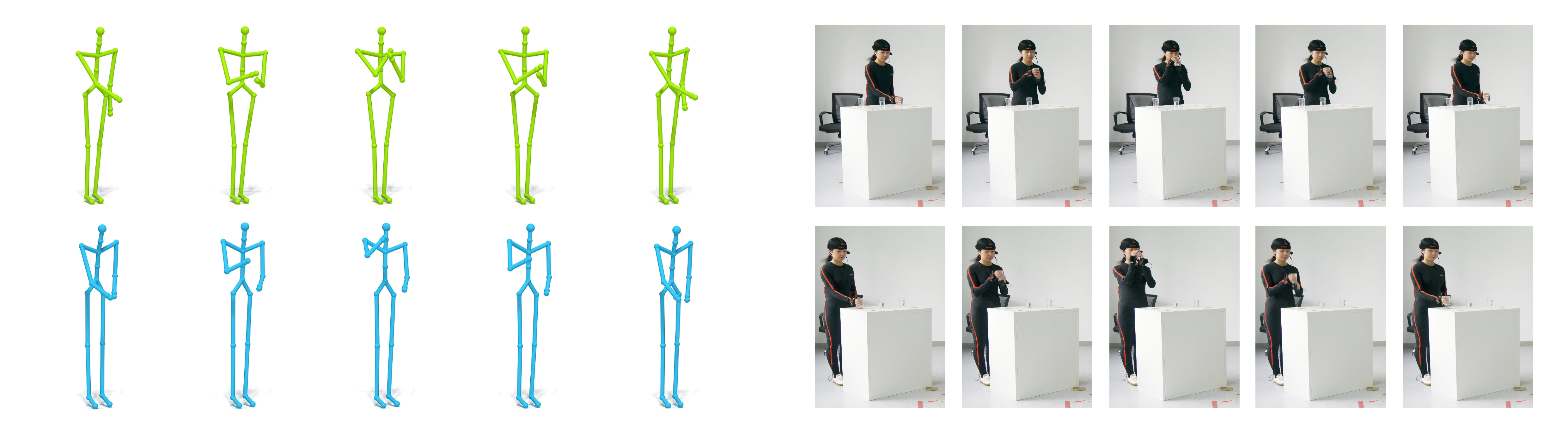}
	\end{center}
	\caption{\hui{Given an unseen motion sequence of drinking from a cup full of water using two hands, we generate a new sequence that drinking from an almost empty cup (the blue skeletons in second row). In the training set, all the subjects drink water using one hand. The corresponding RGB images of the actor are shown in the right for a better illustration.}}
	\label{fig:drinking}
\end{figure*}

\paragraph*{Results.}
Figures~\ref{fig:walking_transfer},~\ref{fig:lifting_transfer}, and~\ref{fig:drinking} show three generated motions by changing the object property values. Please also refer to the supplementary video for more examples. 
When the input is a walking motion on a width path by an unseen subject, we transfer motion to walk on a narrow path, like a catwalk model.
Given a motion sequence of an unseen subject lifting a light box from a table to a closet, we generate a new sequence that looks like the box is too heavy to be lifted up; see Figure~\ref{fig:lifting_transfer}.  

%
%

In Figure~\ref{fig:drinking}, we show a generated sequence that drinking from an empty cup, given an unseen motion sequence drinking from a cup full of water using two hands. 
As the unseen motion is considerably different from the training set, the generated motion deviates from the input. 
However, sometimes it is ambiguous what is the correct motion. 
Note during training, we constrain the synthesized motion conditioned on a target property value to be similar to the motion performed by the same subject of given object property. Multiple options may likely match the desired motion property value. It would be desirable if we could synthesize the one that is most similar to the input motion.

\subsection{User study}
We conducted two user studies. 
The first one is to investigate a human observer's perception on the weight and fragility inference from skeleton sequences. We considered both the 3D skeletons captured and 2D skeletons extracted from videos.
The second user study is conducted to evaluate the property-aware motion transfer on the sitting and walking sequences.

\paragraph*{The first user study.}
In the study, a test consisted of watching a video of skeletal motion of an actor lifting a box or moving a bowl, 
then predicting the unseen object's property by choosing an answer from multiple choices.
For LiftingBox sequence, six choices were provided: 0, 5, 10, 15, 20, and 25kg. For MovingBowl sequence, three choices were provided: empty, fully filled with rice, and fully filled with water. There were a total of 12 tests. To help answering the questions, 4 demos with correct answers were played before the tests started. These motion samples were randomly chosen from the testing group. Each video was about \qian{3--6} seconds long. All participants had full control over these videos, e.g., start, pause, stop and navigate in time, etc.
A total of 60 participants were recruited.
Each participant did the user study twice. The first time they predicted the weight from videos of rendered 3D skeletons, and the second time they predicted the weight from 2D video skeletons. 
\hui{Note that 2D video skeletons have large missing parts in some frames due to the occlusions introduced by human body shape or the objects, while the rendered ones have much fewer occlusion cases caused by bones.}
These skeletons were drawn with the same color encoding.
The total study time for each participant was around 10 minutes.

Figure~\ref{fig:weight_fragility} (left) shows the average predicted weights by users and our model for boxes of different physical weights. The estimated weights by our model using 3D skeletons as input are much closer to the physical ground truth than other settings. 
Note that our reported human performance is slightly lower than that reported in Runeson and Frykholm's work~\shortcite{Runeson1981}. A possible reason is that a smaller weight step (5kg) and more weight classes (6) were used in our user study.
Figure~\ref{fig:weight_fragility} (right) displays the F1 scores of user study and our model on the fragility inference. Note it is challenging to distinguish an empty bowl from a bowl full of rice, but still, our model outperformed on both 2D and 3D skeletal cases. 

\begin{figure}[t]
	\begin{center}
		\includegraphics[width=0.49\linewidth]{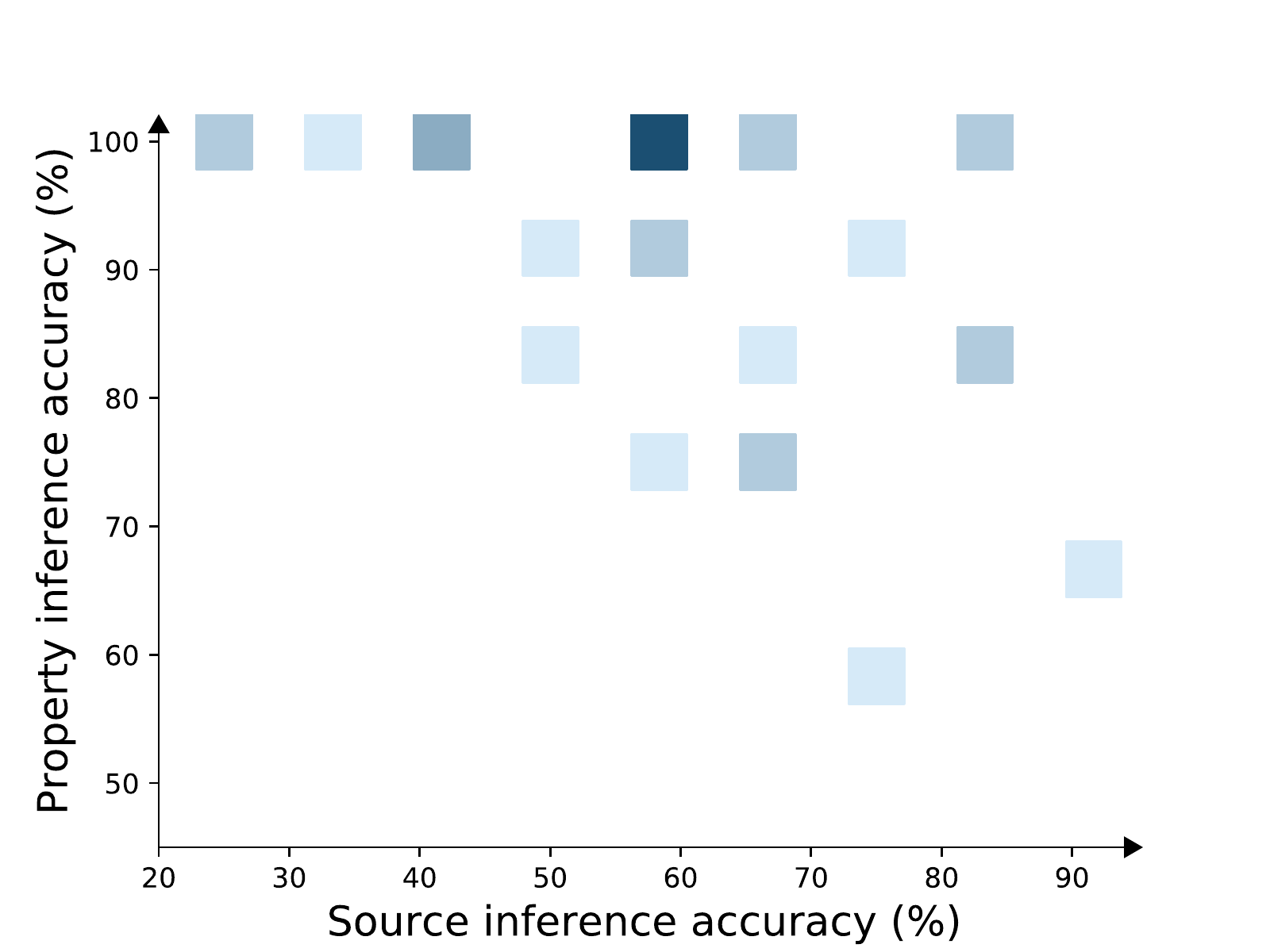}
		\includegraphics[width=0.49\linewidth]{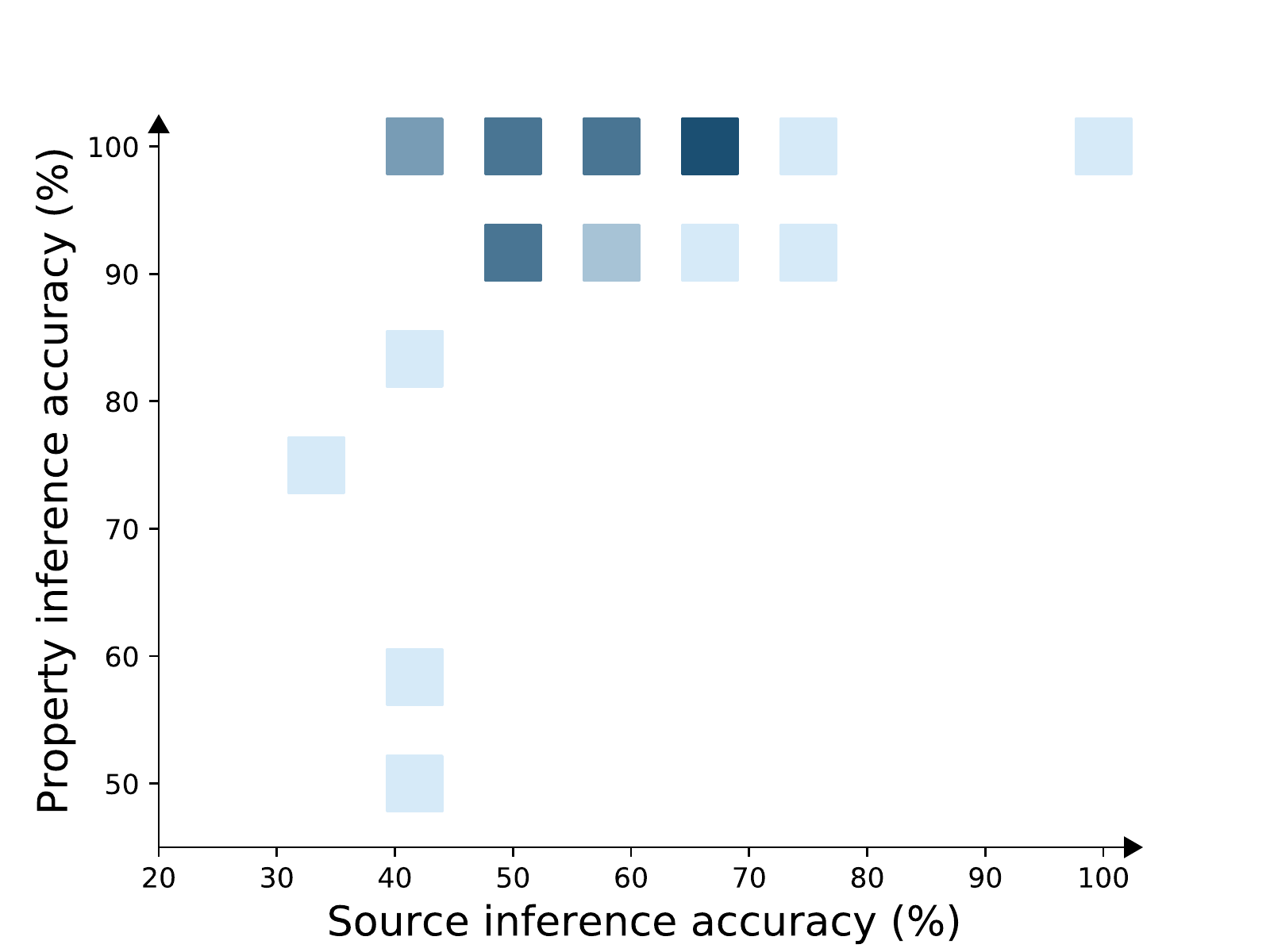}
	\end{center}
	\caption{The scatter map of participants' accuracy (\%) on guess the motion's source (synthesized or captured), and on the object property inference. Left: sitting; right: walking. Shades of blue indicate the number of participants, darker being higher.} 
	\label{fig:user_study_transfer}
\end{figure}

\paragraph*{The second user study.}
A total of 60 participants were recruited and divided into two groups, watching the sitting and walking sequences, respectively. 
Every participant did 12 tests, and 4 demos with correct answers were played before the tests started.
A test contained two parts.
The first task is to judge if the given motion was synthesized or captured. 
The second task is to select the associated object property of the given motion, while only 2 choices were provided. For example, to select the path being walked on was wide or narrow, or the chair being sit on was soft or hard.
Other settings are similar to the first user study. 
Figure~\ref{fig:user_study_transfer} shows the performance of participants on motion source and object property inference. The lightness of a square encodes the number of participants with a particular inference accuracy, the darker the higher. 
For majority participants, the source inference accuracy is about 60\%, while the property inference accuracy is above 90\%, indicating that our synthesized motions are quite close to real captured ones.

\section{Conclusions and Future Work}

The primary goal of this work is to study human interaction motions represented by skeleton sequences, and investigate whether and how well a machine can learn to infer the properties of unseen interacting objects, and to what extent we can have control on the synthesis of motions with target object properties.
\hui{We have built up a large multi-modal dataset for such object property inference from fine-grained human interaction motions with 4,000+ samples, which consist of 100 participants performing 8 different tasks, and thus related to 8 different object properties.} 

Using 3D skeleton sequences alone, we have learned to infer the properties of interacting objects by treating it as a classification problem, and evaluated our trained model in various settings. The collected 3D skeleton sequences allows data-driven learning, and help achieve better inference accuracy in comparison with using other data sources or even human observers.
We have presented a network to disentangle object property from the motion. The disentangling, in turn, allows the synthesis of modified motion with a target object property. This control over the actions enriches the dataset on one hand, and optimizes the specific animation of particular individuals on the other.  

\paragraph*{Limitations.}
Due to the design, our target problem is limited in the defined scenarios with pre-defined human motions and object properties. The inference and transfer tasks are solved separately, while exploiting features extracted during inference to guide the synthesis part might be possibly better.

The object property classifier is evaluated on eight types of motions, and the accuracy is not that high. We focus on the intra-class characteristics for the object property inference, but it might be better to address action recognition and object property inference altogether, as the action types provide more global content information.

The object property-aware motion transfer employs an encoder-decoder structure with 1D convolution layers, which might not fully capture the spatial-temporal information of more human motions, in particular, the complex ones. More advanced network structures, such as STRNN~\cite{WangMonifold2019}, could be used to better transfer in-between independent actions.

\paragraph*{Future work.} 
Exciting research directions lay ahead as we are only starting to exploit the collected motion data. We would make the very large-scale interaction dataset public. We believe that this dataset will stimulate further research, and in the future, we will strive not only to increase the number of samples, but also the types of human-object interactions.
Previous works have shown that some other properties, e.g., size and geometric shape, are quite hard to be estimated from a pantomimed action~\cite{Vaina1995}. To be able to deal with more diverse object properties, we are also considering fusing more visual inputs, e.g., videos and depth sequences, with 3D skeletal motions.

Another promising direction is to discover exactly which parts of the skeleton are critical for the specific object property inference, by considering more sophisticated attention models or computing more advanced skeletal features. Further exploration could also focus on designing new networks that can learn and encode skeletal motions in a learned latent space, instead of being explicitly provided parameterization. It is certainly more exciting if we can directly predict object properties from 2D video inputs with high accuracy using a trained model on 3D skeletal motions, eventually leading to new modes of authoring video sequences.

\begin{acks}
	
This work was supported in parts by \grantsponsor{JCY}{Shenzhen Innovation Program}~ (\grantnum{JCY}{JCYJ20180305125709986}) and \grantsponsor{NSFC}{NSFC}~ (\grantnum{NSFC}{61861130365}, \grantnum{NSFC}{61761146002}).

\end{acks}


------------------
\bibliographystyle{ACM-Reference-Format}
\bibliography{it-ref}       


\begin{thebibliography}{60}


\ifx \showCODEN    \undefined \def \showCODEN     #1{\unskip}     \fi
\ifx \showDOI      \undefined \def \showDOI       #1{#1}\fi
\ifx \showISBNx    \undefined \def \showISBNx     #1{\unskip}     \fi
\ifx \showISBNxiii \undefined \def \showISBNxiii  #1{\unskip}     \fi
\ifx \showISSN     \undefined \def \showISSN      #1{\unskip}     \fi
\ifx \showLCCN     \undefined \def \showLCCN      #1{\unskip}     \fi
\ifx \shownote     \undefined \def \shownote      #1{#1}          \fi
\ifx \showarticletitle \undefined \def \showarticletitle #1{#1}   \fi
\ifx \showURL      \undefined \def \showURL       {\relax}        \fi
\providecommand\bibfield[2]{#2}
\providecommand\bibinfo[2]{#2}
\providecommand\natexlab[1]{#1}
\providecommand\showeprint[2][]{arXiv:#2}

\bibitem[\protect\citeauthoryear{Aberman, Wu, Lischinski, Chen, and
  Cohen-Or}{Aberman et~al\mbox{.}}{2019}]%
        {Aberman2019}
\bibfield{author}{\bibinfo{person}{Kfir Aberman}, \bibinfo{person}{Rundi Wu},
  \bibinfo{person}{Dani Lischinski}, \bibinfo{person}{Baoquan Chen}, {and}
  \bibinfo{person}{Daniel Cohen-Or}.} \bibinfo{year}{2019}\natexlab{}.
\newblock \showarticletitle{Learning Character-Agnostic Motion for Motion
  Retargeting in 2D}.
\newblock \bibinfo{journal}{\emph{ACM Trans. on Graphics (Proc. of SIGGRAPH)}}
  \bibinfo{volume}{38}, \bibinfo{number}{4} (\bibinfo{year}{2019}),
  \bibinfo{pages}{75}.
\newblock
\urldef\tempurl%
\url{https://doi.org/10.1145/3306346.3322999}
\showDOI{\tempurl}


\bibitem[\protect\citeauthoryear{Andriluka, Iqbal, Insafutdinov, Pishchulin,
  Milan, Gall, and Schiele}{Andriluka et~al\mbox{.}}{2018}]%
        {Andriluka2018}
\bibfield{author}{\bibinfo{person}{Mykhaylo Andriluka}, \bibinfo{person}{Umar
  Iqbal}, \bibinfo{person}{Eldar Insafutdinov}, \bibinfo{person}{Leonid
  Pishchulin}, \bibinfo{person}{Anton Milan}, \bibinfo{person}{Juergen Gall},
  {and} \bibinfo{person}{Bernt Schiele}.} \bibinfo{year}{2018}\natexlab{}.
\newblock \showarticletitle{PoseTrack: A Benchmark for Human Pose Estimation
  and Tracking}. In \bibinfo{booktitle}{\emph{CVPR}}.
\newblock
\urldef\tempurl%
\url{https://doi.org/10.1109/CVPR.2018.00542}
\showDOI{\tempurl}


\bibitem[\protect\citeauthoryear{Aristidou, Cohen-Or, Hodgins, Chrysanthou, and
  Shamir}{Aristidou et~al\mbox{.}}{2018}]%
        {Aristidou2018}
\bibfield{author}{\bibinfo{person}{Andreas Aristidou}, \bibinfo{person}{Daniel
  Cohen-Or}, \bibinfo{person}{Jessica~K. Hodgins}, \bibinfo{person}{Yiorgos
  Chrysanthou}, {and} \bibinfo{person}{Ariel Shamir}.}
  \bibinfo{year}{2018}\natexlab{}.
\newblock \showarticletitle{Deep Motifs and Motion Signatures}.
\newblock \bibinfo{journal}{\emph{ACM Trans. on Graphics (Proc. of SIGGRAPH
  Asia)}} \bibinfo{volume}{38}, \bibinfo{number}{6} (\bibinfo{year}{2018}),
  \bibinfo{pages}{187:1--187:13}.
\newblock
\urldef\tempurl%
\url{https://doi.org/10.1145/3272127.3275038}
\showDOI{\tempurl}


\bibitem[\protect\citeauthoryear{Blake and Shiffrar}{Blake and
  Shiffrar}{2007}]%
        {Blake2007}
\bibfield{author}{\bibinfo{person}{Randolph Blake} {and}
  \bibinfo{person}{Maggie Shiffrar}.} \bibinfo{year}{2007}\natexlab{}.
\newblock \showarticletitle{Perception of Human Motion}.
\newblock \bibinfo{journal}{\emph{Annual Review of Psychology}}
  \bibinfo{volume}{58}, \bibinfo{number}{1} (\bibinfo{year}{2007}),
  \bibinfo{pages}{47--73}.
\newblock
\urldef\tempurl%
\url{https://doi.org/10.1146/annurev.psych.57.102904.190152}
\showDOI{\tempurl}


\bibitem[\protect\citeauthoryear{Cao, Simon, Wei, and Sheikh}{Cao
  et~al\mbox{.}}{2017}]%
        {Cao2017}
\bibfield{author}{\bibinfo{person}{Zhe Cao}, \bibinfo{person}{Tomas Simon},
  \bibinfo{person}{Shih-En Wei}, {and} \bibinfo{person}{Yaser Sheikh}.}
  \bibinfo{year}{2017}\natexlab{}.
\newblock \showarticletitle{Realtime Multi-person 2D Pose Estimation Using Part
  Affinity Fields}. In \bibinfo{booktitle}{\emph{CVPR}}.
\newblock
\urldef\tempurl%
\url{https://doi.org/10.1109/CVPR.2017.143}
\showDOI{\tempurl}


\bibitem[\protect\citeauthoryear{Charles, Su, Kaichun, and Guibas}{Charles
  et~al\mbox{.}}{2017}]%
        {Charles2017}
\bibfield{author}{\bibinfo{person}{R.~Qi Charles}, \bibinfo{person}{Hao Su},
  \bibinfo{person}{Mo Kaichun}, {and} \bibinfo{person}{Leonidas~J. Guibas}.}
  \bibinfo{year}{2017}\natexlab{}.
\newblock \showarticletitle{{PointNet}: Deep Learning on Point Sets for 3D
  Classification and Segmentation}. In \bibinfo{booktitle}{\emph{CVPR}}.
\newblock
\urldef\tempurl%
\url{https://doi.org/10.1109/cvpr.2017.16}
\showDOI{\tempurl}


\bibitem[\protect\citeauthoryear{Cho, Van~Merri{\"e}nboer, Gulcehre, Bahdanau,
  Bougares, Schwenk, and Bengio}{Cho et~al\mbox{.}}{2014}]%
        {cho2014learning}
\bibfield{author}{\bibinfo{person}{Kyunghyun Cho}, \bibinfo{person}{Bart
  Van~Merri{\"e}nboer}, \bibinfo{person}{Caglar Gulcehre},
  \bibinfo{person}{Dzmitry Bahdanau}, \bibinfo{person}{Fethi Bougares},
  \bibinfo{person}{Holger Schwenk}, {and} \bibinfo{person}{Yoshua Bengio}.}
  \bibinfo{year}{2014}\natexlab{}.
\newblock \showarticletitle{Learning phrase representations using RNN
  encoder-decoder for statistical machine translation}.
\newblock \bibinfo{journal}{\emph{arXiv preprint arXiv:1406.1078}}
  (\bibinfo{year}{2014}).
\newblock


\bibitem[\protect\citeauthoryear{CMU}{CMU}{2018}]%
        {CMU2018}
\bibfield{author}{\bibinfo{person}{CMU}.} \bibinfo{year}{2018}\natexlab{}.
\newblock \bibinfo{title}{Carnegie Mellon University MoCap Database}.
\newblock
\newblock


\bibitem[\protect\citeauthoryear{Davis and Gao}{Davis and Gao}{2003}]%
        {Davis2003}
\bibfield{author}{\bibinfo{person}{James~W. Davis} {and} \bibinfo{person}{Hui
  Gao}.} \bibinfo{year}{2003}\natexlab{}.
\newblock \showarticletitle{Recognizing human action efforts: an adaptive
  three-mode {PCA} framework}. In \bibinfo{booktitle}{\emph{ICCV}}.
\newblock


\bibitem[\protect\citeauthoryear{de~C.~Hamilton, Joyce, Flanagan, Frith, and
  Wolpert}{de~C.~Hamilton et~al\mbox{.}}{2005}]%
        {C.Hamilton2005}
\bibfield{author}{\bibinfo{person}{A.~F. de C.~Hamilton},
  \bibinfo{person}{D.~W. Joyce}, \bibinfo{person}{J.~R. Flanagan},
  \bibinfo{person}{C.~D. Frith}, {and} \bibinfo{person}{D.~M. Wolpert}.}
  \bibinfo{year}{2005}\natexlab{}.
\newblock \showarticletitle{Kinematic cues in perceptual weight judgement and
  their origins in box lifting}.
\newblock \bibinfo{journal}{\emph{Psychological Research}}
  \bibinfo{volume}{71}, \bibinfo{number}{1} (\bibinfo{year}{2005}),
  \bibinfo{pages}{13--21}.
\newblock
\urldef\tempurl%
\url{https://doi.org/10.1007/s00426-005-0032-4}
\showDOI{\tempurl}


\bibitem[\protect\citeauthoryear{Gkioxari, Girshick, Doll\'{a}r, and
  He}{Gkioxari et~al\mbox{.}}{2018}]%
        {Gkioxari2018}
\bibfield{author}{\bibinfo{person}{Georgia Gkioxari}, \bibinfo{person}{Ross
  Girshick}, \bibinfo{person}{Piotr Doll\'{a}r}, {and} \bibinfo{person}{Kaiming
  He}.} \bibinfo{year}{2018}\natexlab{}.
\newblock \showarticletitle{Detecting and Recognizing Human-Object
  Intaractions}.
\newblock \bibinfo{journal}{\emph{CVPR}} (\bibinfo{year}{2018}).
\newblock
\urldef\tempurl%
\url{https://doi.org/10.1109/CVPR.2018.00872}
\showDOI{\tempurl}


\bibitem[\protect\citeauthoryear{Grabner, Gall, and Gool}{Grabner
  et~al\mbox{.}}{2011}]%
        {Grabner2011}
\bibfield{author}{\bibinfo{person}{Helmut Grabner}, \bibinfo{person}{Juergen
  Gall}, {and} \bibinfo{person}{Luc~Van Gool}.}
  \bibinfo{year}{2011}\natexlab{}.
\newblock \showarticletitle{What makes a chair a chair?}. In
  \bibinfo{booktitle}{\emph{CVPR}}.
\newblock
\urldef\tempurl%
\url{https://doi.org/10.1109/CVPR.2011.5995327}
\showDOI{\tempurl}


\bibitem[\protect\citeauthoryear{Gui, Wang, Liang, and Moura}{Gui
  et~al\mbox{.}}{2018}]%
        {Gui2018}
\bibfield{author}{\bibinfo{person}{Liang-Yan Gui}, \bibinfo{person}{Yu-Xiong
  Wang}, \bibinfo{person}{Xiaodan Liang}, {and} \bibinfo{person}{Jos{\'{e}}
  M.~F. Moura}.} \bibinfo{year}{2018}\natexlab{}.
\newblock \showarticletitle{Adversarial Geometry-Aware Human Motion
  Prediction}.
\newblock In \bibinfo{booktitle}{\emph{ECCV}}. \bibinfo{pages}{823--842}.
\newblock
\urldef\tempurl%
\url{https://doi.org/10.1007/978-3-030-01225-0_48}
\showDOI{\tempurl}


\bibitem[\protect\citeauthoryear{Gupta and Davis}{Gupta and Davis}{2007}]%
        {Gupta2007}
\bibfield{author}{\bibinfo{person}{Abhinav Gupta} {and}
  \bibinfo{person}{Larry~S. Davis}.} \bibinfo{year}{2007}\natexlab{}.
\newblock \showarticletitle{Objects in Action: An Approach for Combining Action
  Understanding and Object Perception}. In \bibinfo{booktitle}{\emph{CVPR}}.
\newblock
\urldef\tempurl%
\url{https://doi.org/10.1109/CVPR.2007.383331}
\showDOI{\tempurl}


\bibitem[\protect\citeauthoryear{{Hadsell}, {Chopra}, and {LeCun}}{{Hadsell}
  et~al\mbox{.}}{2006}]%
        {Hadsell2006}
\bibfield{author}{\bibinfo{person}{R. {Hadsell}}, \bibinfo{person}{S.
  {Chopra}}, {and} \bibinfo{person}{Y. {LeCun}}.}
  \bibinfo{year}{2006}\natexlab{}.
\newblock \showarticletitle{Dimensionality Reduction by Learning an Invariant
  Mapping}. In \bibinfo{booktitle}{\emph{CVPR}}, Vol.~\bibinfo{volume}{2}.
  \bibinfo{pages}{1735--1742}.
\newblock
\urldef\tempurl%
\url{https://doi.org/10.1109/CVPR.2006.100}
\showDOI{\tempurl}


\bibitem[\protect\citeauthoryear{Ho, Komura, and Tai}{Ho et~al\mbox{.}}{2010}]%
        {Ho2010}
\bibfield{author}{\bibinfo{person}{Edmond S.~L. Ho}, \bibinfo{person}{Taku
  Komura}, {and} \bibinfo{person}{Chiew-Lan Tai}.}
  \bibinfo{year}{2010}\natexlab{}.
\newblock \showarticletitle{Spatial Relationship Preserving Character Motion
  Adaptation}.
\newblock \bibinfo{journal}{\emph{ACM Trans. on Graphics (Proc. of SIGGRAPH)}}
  \bibinfo{volume}{29}, \bibinfo{number}{4}, Article \bibinfo{articleno}{33}
  (\bibinfo{date}{July} \bibinfo{year}{2010}).
\newblock
\urldef\tempurl%
\url{https://doi.org/10.1145/1778765.1778770}
\showDOI{\tempurl}


\bibitem[\protect\citeauthoryear{Holden, Saito, and Komura}{Holden
  et~al\mbox{.}}{2016}]%
        {Holden2016}
\bibfield{author}{\bibinfo{person}{Daniel Holden}, \bibinfo{person}{Jun Saito},
  {and} \bibinfo{person}{Taku Komura}.} \bibinfo{year}{2016}\natexlab{}.
\newblock \showarticletitle{A deep learning framework for character motion
  synthesis and editing}.
\newblock \bibinfo{journal}{\emph{ACM Trans. on Graphics (Proc. of SIGGRAPH)}}
  \bibinfo{volume}{35}, \bibinfo{number}{4} (\bibinfo{year}{2016}),
  \bibinfo{pages}{1--11}.
\newblock
\urldef\tempurl%
\url{https://doi.org/10.1145/2897824.2925975}
\showDOI{\tempurl}


\bibitem[\protect\citeauthoryear{Hsu, Pulli, and Popovi\'{c}}{Hsu
  et~al\mbox{.}}{2005}]%
        {Hsu2005}
\bibfield{author}{\bibinfo{person}{Eugene Hsu}, \bibinfo{person}{Kari Pulli},
  {and} \bibinfo{person}{Jovan Popovi\'{c}}.} \bibinfo{year}{2005}\natexlab{}.
\newblock \showarticletitle{Style Translation for Human Motion}.
\newblock \bibinfo{journal}{\emph{ACM Trans. on Graphics}}
  \bibinfo{volume}{24}, \bibinfo{number}{3} (\bibinfo{date}{July}
  \bibinfo{year}{2005}), \bibinfo{pages}{1082–1089}.
\newblock
\urldef\tempurl%
\url{https://doi.org/10.1145/1073204.1073315}
\showDOI{\tempurl}


\bibitem[\protect\citeauthoryear{Hu, Savva, and van Kaick}{Hu
  et~al\mbox{.}}{2018a}]%
        {Hu2018}
\bibfield{author}{\bibinfo{person}{Ruizhen Hu}, \bibinfo{person}{Manolis
  Savva}, {and} \bibinfo{person}{Oliver van Kaick}.}
  \bibinfo{year}{2018}\natexlab{a}.
\newblock \showarticletitle{Functionality Representations and Applications for
  Shape Analysis}.
\newblock \bibinfo{journal}{\emph{Computer Graphics Forum (Proc. of
  Eurographics)}} \bibinfo{volume}{37}, \bibinfo{number}{2}
  (\bibinfo{year}{2018}), \bibinfo{pages}{603--624}.
\newblock
\urldef\tempurl%
\url{https://doi.org/10.1111/cgf.13385}
\showDOI{\tempurl}


\bibitem[\protect\citeauthoryear{Hu, Yan, Zhang, Kaick, Shamir, Zhang, and
  Huang}{Hu et~al\mbox{.}}{2018b}]%
        {Hu2018a}
\bibfield{author}{\bibinfo{person}{Ruizhen Hu}, \bibinfo{person}{Zihao Yan},
  \bibinfo{person}{Jingwen Zhang}, \bibinfo{person}{Oliver~Van Kaick},
  \bibinfo{person}{Ariel Shamir}, \bibinfo{person}{Hao Zhang}, {and}
  \bibinfo{person}{Hui Huang}.} \bibinfo{year}{2018}\natexlab{b}.
\newblock \showarticletitle{Predictive and generative neural networks for
  object functionality}.
\newblock \bibinfo{journal}{\emph{ACM Trans. on Graphics (Proc. of SIGGRAPH)}}
  \bibinfo{volume}{37}, \bibinfo{number}{4} (\bibinfo{year}{2018}),
  \bibinfo{pages}{1--13}.
\newblock
\urldef\tempurl%
\url{https://doi.org/10.1145/3197517.3201287}
\showDOI{\tempurl}


\bibitem[\protect\citeauthoryear{Insafutdinov, Pishchulin, Andres, Andriluka,
  and Schiele}{Insafutdinov et~al\mbox{.}}{2016}]%
        {Insafutdinov2016}
\bibfield{author}{\bibinfo{person}{Eldar Insafutdinov}, \bibinfo{person}{Leonid
  Pishchulin}, \bibinfo{person}{Bjoern Andres}, \bibinfo{person}{Mykhaylo
  Andriluka}, {and} \bibinfo{person}{Bernt Schiele}.}
  \bibinfo{year}{2016}\natexlab{}.
\newblock \showarticletitle{{DeeperCut}: A Deeper, Stronger, and Faster
  Multi-person Pose Estimation Model}. In \bibinfo{booktitle}{\emph{ECCV}}.
\newblock
\urldef\tempurl%
\url{https://doi.org/10.1007/978-3-319-46466-4_3}
\showDOI{\tempurl}


\bibitem[\protect\citeauthoryear{Jiang, Koppula, and Saxena}{Jiang
  et~al\mbox{.}}{2013}]%
        {Jiang2013}
\bibfield{author}{\bibinfo{person}{Yun Jiang}, \bibinfo{person}{Hema Koppula},
  {and} \bibinfo{person}{Ashutosh Saxena}.} \bibinfo{year}{2013}\natexlab{}.
\newblock \showarticletitle{Hallucinated Humans as the Hidden Context for
  Labeling 3D Scenes}. In \bibinfo{booktitle}{\emph{CVPR}}.
  \bibinfo{pages}{2993--3000}.
\newblock
\urldef\tempurl%
\url{https://doi.org/10.1109/CVPR.2013.385}
\showDOI{\tempurl}


\bibitem[\protect\citeauthoryear{Jiang, Koppula, and Saxena}{Jiang
  et~al\mbox{.}}{2016}]%
        {Jiang2016}
\bibfield{author}{\bibinfo{person}{Yun Jiang}, \bibinfo{person}{Hema Koppula},
  {and} \bibinfo{person}{Ashutosh Saxena}.} \bibinfo{year}{2016}\natexlab{}.
\newblock \showarticletitle{Modeling 3D Environments through Hidden Human
  Context}.
\newblock \bibinfo{journal}{\emph{IEEE Trans. Pattern Anal. Mach. Intell.}}
  \bibinfo{volume}{38}, \bibinfo{number}{10} (\bibinfo{year}{2016}),
  \bibinfo{pages}{2040--2053}.
\newblock
\urldef\tempurl%
\url{https://doi.org/10.1109/TPAMI.2015.2501811}
\showDOI{\tempurl}


\bibitem[\protect\citeauthoryear{Kanazawa, Black, Jacobs, and Malik}{Kanazawa
  et~al\mbox{.}}{2018}]%
        {Kanazawa2018}
\bibfield{author}{\bibinfo{person}{Angjoo Kanazawa},
  \bibinfo{person}{Michael~J. Black}, \bibinfo{person}{David~W. Jacobs}, {and}
  \bibinfo{person}{Jitendra Malik}.} \bibinfo{year}{2018}\natexlab{}.
\newblock \showarticletitle{End-to-End Recovery of Human Shape and Pose}. In
  \bibinfo{booktitle}{\emph{CVPR}}.
\newblock
\urldef\tempurl%
\url{https://doi.org/10.1109/CVPR.2018.00744}
\showDOI{\tempurl}


\bibitem[\protect\citeauthoryear{Kang and Lee}{Kang and Lee}{2017}]%
        {Kang2017}
\bibfield{author}{\bibinfo{person}{Changgu Kang} {and}
  \bibinfo{person}{Sung-Hee Lee}.} \bibinfo{year}{2017}\natexlab{}.
\newblock \showarticletitle{Scene reconstruction and analysis from motion}.
\newblock \bibinfo{journal}{\emph{Graphical Models}}  \bibinfo{volume}{94}
  (\bibinfo{year}{2017}), \bibinfo{pages}{25--37}.
\newblock
\urldef\tempurl%
\url{https://doi.org/10.1016/j.gmod.2017.10.002}
\showDOI{\tempurl}


\bibitem[\protect\citeauthoryear{Kato, Li, and Gupta}{Kato
  et~al\mbox{.}}{2018}]%
        {Kato2018}
\bibfield{author}{\bibinfo{person}{Keizo Kato}, \bibinfo{person}{Yin Li}, {and}
  \bibinfo{person}{Abhinav Gupta}.} \bibinfo{year}{2018}\natexlab{}.
\newblock \showarticletitle{Compositional Learning for Human Object
  Interaction}. In \bibinfo{booktitle}{\emph{ECCV}}.
\newblock
\urldef\tempurl%
\url{https://doi.org/10.1007/978-3-030-01264-9_15}
\showDOI{\tempurl}


\bibitem[\protect\citeauthoryear{Ke, Bennamoun, An, Sohel, and Boussaid}{Ke
  et~al\mbox{.}}{2017}]%
        {Ke2017}
\bibfield{author}{\bibinfo{person}{Qiuhong Ke}, \bibinfo{person}{Mohammed
  Bennamoun}, \bibinfo{person}{Senjian An}, \bibinfo{person}{Ferdous Sohel},
  {and} \bibinfo{person}{Farid Boussaid}.} \bibinfo{year}{2017}\natexlab{}.
\newblock \showarticletitle{A New Representation of Skeleton Sequences for 3D
  Action Recognition}. In \bibinfo{booktitle}{\emph{CVPR}}.
\newblock
\urldef\tempurl%
\url{https://doi.org/10.1109/CVPR.2017.486}
\showDOI{\tempurl}


\bibitem[\protect\citeauthoryear{Kim, Chaudhuri, Guibas, and Funkhouser}{Kim
  et~al\mbox{.}}{2014}]%
        {Kim2014}
\bibfield{author}{\bibinfo{person}{Vladimir~G. Kim},
  \bibinfo{person}{Siddhartha Chaudhuri}, \bibinfo{person}{Leonidas Guibas},
  {and} \bibinfo{person}{Thomas Funkhouser}.} \bibinfo{year}{2014}\natexlab{}.
\newblock \showarticletitle{Shape2Pose: human-centric shape analysis}.
\newblock \bibinfo{journal}{\emph{ACM Trans. on Graphics (Proc. of SIGGRAPH)}}
  \bibinfo{volume}{33}, \bibinfo{number}{4} (\bibinfo{year}{2014}),
  \bibinfo{pages}{1--12}.
\newblock
\urldef\tempurl%
\url{https://doi.org/10.1145/2601097.2601117}
\showDOI{\tempurl}


\bibitem[\protect\citeauthoryear{Li, Zhong, Xie, and Pu}{Li
  et~al\mbox{.}}{2018}]%
        {Li2018}
\bibfield{author}{\bibinfo{person}{Chao Li}, \bibinfo{person}{Qiaoyong Zhong},
  \bibinfo{person}{Di Xie}, {and} \bibinfo{person}{Shiliang Pu}.}
  \bibinfo{year}{2018}\natexlab{}.
\newblock \showarticletitle{Co-occurrence Feature Learning from Skeleton Data
  for Action Recognition and Detection with Hierarchical Aggregation}. In
  \bibinfo{booktitle}{\emph{Proc. Int. Joint Conf. on Artificial
  Intelligence}}. \bibinfo{pages}{786–792}.
\newblock
\urldef\tempurl%
\url{https://doi.org/10.5555/3304415.3304527}
\showDOI{\tempurl}


\bibitem[\protect\citeauthoryear{Li, Liu, Kim, Wang, Yang, and Kautz}{Li
  et~al\mbox{.}}{2019}]%
        {Li2019}
\bibfield{author}{\bibinfo{person}{Xueting Li}, \bibinfo{person}{Sifei Liu},
  \bibinfo{person}{Kihwan Kim}, \bibinfo{person}{Xiaolong Wang},
  \bibinfo{person}{Ming-Hsuan Yang}, {and} \bibinfo{person}{Jan Kautz}.}
  \bibinfo{year}{2019}\natexlab{}.
\newblock \showarticletitle{Putting Humans in a Scene: Learning Affordance in
  3D Indoor Environments}. In \bibinfo{booktitle}{\emph{CVPR}}.
\newblock
\urldef\tempurl%
\url{https://doi.org/10.1109/CVPR.2019.01265}
\showDOI{\tempurl}


\bibitem[\protect\citeauthoryear{Liu, Hu, Li, Song, and Liu}{Liu
  et~al\mbox{.}}{2017a}]%
        {Liu2017a}
\bibfield{author}{\bibinfo{person}{Chunhui Liu}, \bibinfo{person}{Yueyu Hu},
  \bibinfo{person}{Yanghao Li}, \bibinfo{person}{Sijie Song}, {and}
  \bibinfo{person}{Jiaying Liu}.} \bibinfo{year}{2017}\natexlab{a}.
\newblock \showarticletitle{PKU-MMD: A Large Scale Benchmark for Continuous
  Multi-Modal Human Action Understanding}.
\newblock \bibinfo{journal}{\emph{arXiv preprint arXiv:1703.07475}}
  (\bibinfo{year}{2017}).
\newblock


\bibitem[\protect\citeauthoryear{Liu, Shahroudy, Xu, and Wang}{Liu
  et~al\mbox{.}}{2016}]%
        {Liu2016}
\bibfield{author}{\bibinfo{person}{Jun Liu}, \bibinfo{person}{Amir Shahroudy},
  \bibinfo{person}{Dong Xu}, {and} \bibinfo{person}{Gang Wang}.}
  \bibinfo{year}{2016}\natexlab{}.
\newblock \showarticletitle{Spatio-Temporal {LSTM} with Trust Gates for 3D
  Human Action Recognition}. In \bibinfo{booktitle}{\emph{ECCV}}.
\newblock
\urldef\tempurl%
\url{https://doi.org/10.1007/978-3-319-46487-9_50}
\showDOI{\tempurl}


\bibitem[\protect\citeauthoryear{Liu, Wang, Hu, Duan, and Kot}{Liu
  et~al\mbox{.}}{2017b}]%
        {Liu2017}
\bibfield{author}{\bibinfo{person}{Jun Liu}, \bibinfo{person}{Gang Wang},
  \bibinfo{person}{Ping Hu}, \bibinfo{person}{Ling-Yu Duan}, {and}
  \bibinfo{person}{Alex~C. Kot}.} \bibinfo{year}{2017}\natexlab{b}.
\newblock \showarticletitle{Global Context-Aware Attention {LSTM} Networks for
  3D Action Recognition}. In \bibinfo{booktitle}{\emph{CVPR}}.
\newblock
\urldef\tempurl%
\url{https://doi.org/10.1109/CVPR.2017.391}
\showDOI{\tempurl}


\bibitem[\protect\citeauthoryear{Lo~Presti and La~Cascia}{Lo~Presti and
  La~Cascia}{2016}]%
        {LoPresti2016}
\bibfield{author}{\bibinfo{person}{Liliana Lo~Presti} {and}
  \bibinfo{person}{Marco La~Cascia}.} \bibinfo{year}{2016}\natexlab{}.
\newblock \showarticletitle{3D Skeleton-based Human Action Classification}.
\newblock \bibinfo{journal}{\emph{Pattern Recogn.}} \bibinfo{volume}{53},
  \bibinfo{number}{C} (\bibinfo{date}{May} \bibinfo{year}{2016}),
  \bibinfo{pages}{130--147}.
\newblock
\urldef\tempurl%
\url{https://doi.org/10.1016/j.patcog.2015.11.019}
\showDOI{\tempurl}


\bibitem[\protect\citeauthoryear{Mehta, Sridhar, Sotnychenko, Rhodin, Shafiei,
  Seidel, Xu, Casas, and Theobalt}{Mehta et~al\mbox{.}}{2017}]%
        {Mehta2017}
\bibfield{author}{\bibinfo{person}{Dushyant Mehta}, \bibinfo{person}{Srinath
  Sridhar}, \bibinfo{person}{Oleksandr Sotnychenko}, \bibinfo{person}{Helge
  Rhodin}, \bibinfo{person}{Mohammad Shafiei}, \bibinfo{person}{Hans-Peter
  Seidel}, \bibinfo{person}{Weipeng Xu}, \bibinfo{person}{Dan Casas}, {and}
  \bibinfo{person}{Christian Theobalt}.} \bibinfo{year}{2017}\natexlab{}.
\newblock \showarticletitle{{VNect: }real-time 3D human pose estimation with a
  single RGB camera}.
\newblock \bibinfo{journal}{\emph{ACM Trans. on Graphics (Proc. of SIGGRAPH)}}
  \bibinfo{volume}{36}, \bibinfo{number}{4} (\bibinfo{year}{2017}),
  \bibinfo{pages}{1--14}.
\newblock
\urldef\tempurl%
\url{https://doi.org/10.1145/3072959.3073596}
\showDOI{\tempurl}


\bibitem[\protect\citeauthoryear{Monszpart, Guerrero, Ceylan, Yumer, and
  Mitra}{Monszpart et~al\mbox{.}}{2019}]%
        {Monszpart2018}
\bibfield{author}{\bibinfo{person}{Aron Monszpart}, \bibinfo{person}{Paul
  Guerrero}, \bibinfo{person}{Duygu Ceylan}, \bibinfo{person}{Ersin Yumer},
  {and} \bibinfo{person}{Niloy~J. Mitra}.} \bibinfo{year}{2019}\natexlab{}.
\newblock \showarticletitle{iMapper: Interaction-guided Joint Scene and Human
  Motion Mapping from Monocular Videos}.
\newblock \bibinfo{journal}{\emph{ACM Trans. on Graphics (Proc. of SIGGRAPH)}}
  \bibinfo{volume}{38}, \bibinfo{number}{4} (\bibinfo{date}{July}
  \bibinfo{year}{2019}).
\newblock
\urldef\tempurl%
\url{https://doi.org/10.1145/3306346.3322961}
\showDOI{\tempurl}


\bibitem[\protect\citeauthoryear{Newell, Yang, and Deng}{Newell
  et~al\mbox{.}}{2016}]%
        {Newell2016}
\bibfield{author}{\bibinfo{person}{Alejandro Newell}, \bibinfo{person}{Kaiyu
  Yang}, {and} \bibinfo{person}{Jia Deng}.} \bibinfo{year}{2016}\natexlab{}.
\newblock \showarticletitle{Stacked Hourglass Networks for Human Pose
  Estimation}. In \bibinfo{booktitle}{\emph{ECCV}}.
\newblock
\urldef\tempurl%
\url{https://doi.org/10.1007/978-3-319-46484-8_29}
\showDOI{\tempurl}


\bibitem[\protect\citeauthoryear{Pavlakos, Zhou, and Daniilidis}{Pavlakos
  et~al\mbox{.}}{2018}]%
        {Pavlakos2018}
\bibfield{author}{\bibinfo{person}{Georgios Pavlakos}, \bibinfo{person}{Xiaowei
  Zhou}, {and} \bibinfo{person}{Kostas Daniilidis}.}
  \bibinfo{year}{2018}\natexlab{}.
\newblock \showarticletitle{Ordinal Depth Supervision for 3D Human Pose
  Estimation}. In \bibinfo{booktitle}{\emph{CVPR}}.
\newblock
\urldef\tempurl%
\url{https://doi.org/10.1109/CVPR.2018.00763}
\showDOI{\tempurl}


\bibitem[\protect\citeauthoryear{Podda, Ansuini, Vastano, Cavallo, and
  Becchio}{Podda et~al\mbox{.}}{2017}]%
        {Podda2017}
\bibfield{author}{\bibinfo{person}{Jessica Podda}, \bibinfo{person}{Caterina
  Ansuini}, \bibinfo{person}{Roberta Vastano}, \bibinfo{person}{Andrea
  Cavallo}, {and} \bibinfo{person}{Cristina Becchio}.}
  \bibinfo{year}{2017}\natexlab{}.
\newblock \showarticletitle{The heaviness of invisible objects: Predictive
  weight judgments from observed real and pantomimed grasps}.
\newblock \bibinfo{journal}{\emph{Cognition}}  \bibinfo{volume}{168}
  (\bibinfo{year}{2017}), \bibinfo{pages}{140--145}.
\newblock
\urldef\tempurl%
\url{https://doi.org/10.1016/j.cognition.2017.06.023}
\showDOI{\tempurl}


\bibitem[\protect\citeauthoryear{Runeson and Frykholm}{Runeson and
  Frykholm}{1981}]%
        {Runeson1981}
\bibfield{author}{\bibinfo{person}{Sverker Runeson} {and}
  \bibinfo{person}{Gunilla Frykholm}.} \bibinfo{year}{1981}\natexlab{}.
\newblock \showarticletitle{Visual perception of lifted weight}.
\newblock \bibinfo{journal}{\emph{Journal of Experimental Psychology: Human
  Perception and Performance}} \bibinfo{volume}{7}, \bibinfo{number}{4}
  (\bibinfo{year}{1981}), \bibinfo{pages}{733}.
\newblock
\urldef\tempurl%
\url{https://doi.org/10.1037/0096-1523.7.4.733}
\showDOI{\tempurl}


\bibitem[\protect\citeauthoryear{Rıza, Neverova, and Kokkinos}{Rıza
  et~al\mbox{.}}{2018}]%
        {Riza2018}
\bibfield{author}{\bibinfo{person}{Alp~G\"uler Rıza}, \bibinfo{person}{Natalia
  Neverova}, {and} \bibinfo{person}{Iasonas Kokkinos}.}
  \bibinfo{year}{2018}\natexlab{}.
\newblock \showarticletitle{DensePose: Dense Human Pose Estimation in the
  Wild}. In \bibinfo{booktitle}{\emph{CVPR}}.
\newblock
\urldef\tempurl%
\url{https://doi.org/10.1109/CVPR.2018.00762}
\showDOI{\tempurl}


\bibitem[\protect\citeauthoryear{Savva, Chang, Hanrahan, Fisher, and
  Nie{\ss}ner}{Savva et~al\mbox{.}}{2014}]%
        {Savva2014}
\bibfield{author}{\bibinfo{person}{Manolis Savva}, \bibinfo{person}{Angel~X.
  Chang}, \bibinfo{person}{Pat Hanrahan}, \bibinfo{person}{Matthew Fisher},
  {and} \bibinfo{person}{Matthias Nie{\ss}ner}.}
  \bibinfo{year}{2014}\natexlab{}.
\newblock \showarticletitle{{SceneGrok}: inferring action maps in 3D
  environments}.
\newblock \bibinfo{journal}{\emph{ACM Trans. on Graphics (Proc. of SIGGRAPH)}}
  \bibinfo{volume}{33}, \bibinfo{number}{6} (\bibinfo{year}{2014}),
  \bibinfo{pages}{1--10}.
\newblock
\urldef\tempurl%
\url{https://doi.org/10.1145/2661229.2661230}
\showDOI{\tempurl}


\bibitem[\protect\citeauthoryear{Schmidt, Paulun, van Assen, and
  Fleming}{Schmidt et~al\mbox{.}}{2017}]%
        {Schmidt2017}
\bibfield{author}{\bibinfo{person}{Filipp Schmidt}, \bibinfo{person}{Vivian~C.
  Paulun}, \bibinfo{person}{Jan Jaap~R. van Assen}, {and}
  \bibinfo{person}{Roland~W. Fleming}.} \bibinfo{year}{2017}\natexlab{}.
\newblock \showarticletitle{Inferring the stiffness of unfamiliar objects from
  optical, shape, and motion cues}.
\newblock \bibinfo{journal}{\emph{Journal of Vision}} \bibinfo{volume}{17},
  \bibinfo{number}{3} (\bibinfo{date}{March} \bibinfo{year}{2017}),
  \bibinfo{pages}{18--18}.
\newblock
\urldef\tempurl%
\url{https://doi.org/10.1167/17.3.18}
\showDOI{\tempurl}


\bibitem[\protect\citeauthoryear{Shahroudy, Liu, Ng, and Wang}{Shahroudy
  et~al\mbox{.}}{2016}]%
        {Shahroudy2016}
\bibfield{author}{\bibinfo{person}{Amir Shahroudy}, \bibinfo{person}{Jun Liu},
  \bibinfo{person}{Tian-Tsong Ng}, {and} \bibinfo{person}{Gang Wang}.}
  \bibinfo{year}{2016}\natexlab{}.
\newblock \showarticletitle{{NTU} {RGB+D:} {A} Large Scale Dataset for 3D Human
  Activity Analysis}. In \bibinfo{booktitle}{\emph{CVPR}}.
\newblock
\urldef\tempurl%
\url{https://doi.org/10.1109/CVPR.2016.115}
\showDOI{\tempurl}


\bibitem[\protect\citeauthoryear{Shen, Yang, Ho, and Shum}{Shen
  et~al\mbox{.}}{2019}]%
        {Shen2019}
\bibfield{author}{\bibinfo{person}{Yijun Shen}, \bibinfo{person}{Longzhi Yang},
  \bibinfo{person}{Edmond S.~L. Ho}, {and} \bibinfo{person}{Hubert P.~H.
  Shum}.} \bibinfo{year}{2019}\natexlab{}.
\newblock \showarticletitle{Interaction-based Human Activity Comparison}.
\newblock \bibinfo{journal}{\emph{IEEE Trans. Visualization \& Computer
  Graphics}} (\bibinfo{year}{2019}), \bibinfo{pages}{1--1}.
\newblock
\urldef\tempurl%
\url{https://doi.org/10.1109/TVCG.2019.2893247}
\showDOI{\tempurl}


\bibitem[\protect\citeauthoryear{Song, Lan, Xing, Zeng, and Liu}{Song
  et~al\mbox{.}}{2017}]%
        {Song2017}
\bibfield{author}{\bibinfo{person}{Sijie Song}, \bibinfo{person}{Cuiling Lan},
  \bibinfo{person}{Junliang Xing}, \bibinfo{person}{Wenjun Zeng}, {and}
  \bibinfo{person}{Jiaying Liu}.} \bibinfo{year}{2017}\natexlab{}.
\newblock \showarticletitle{An End-to-End Spatio-Temporal Attention Model for
  Human Action Recognition from Skeleton Data}. In
  \bibinfo{booktitle}{\emph{AAAI Conf. on Artificial Intelligence}}.
\newblock
\urldef\tempurl%
\url{https://doi.org/10.5555/3298023.3298186}
\showDOI{\tempurl}


\bibitem[\protect\citeauthoryear{Stoffregen and Flynn}{Stoffregen and
  Flynn}{1994}]%
        {Stoffregen1994}
\bibfield{author}{\bibinfo{person}{Thomas~A. Stoffregen} {and}
  \bibinfo{person}{Steven~B. Flynn}.} \bibinfo{year}{1994}\natexlab{}.
\newblock \showarticletitle{Visual Perception of Support-Surface Deformability
  From Human Body Kinematics}.
\newblock \bibinfo{journal}{\emph{Ecological Psychology}} \bibinfo{volume}{6},
  \bibinfo{number}{1} (\bibinfo{year}{1994}), \bibinfo{pages}{33--64}.
\newblock
\urldef\tempurl%
\url{https://doi.org/10.1207/s15326969eco0601\_2}
\showDOI{\tempurl}


\bibitem[\protect\citeauthoryear{Tekin, Rozantsev, Lepetit, and Fua}{Tekin
  et~al\mbox{.}}{2016}]%
        {Tekin2016}
\bibfield{author}{\bibinfo{person}{Bugra Tekin}, \bibinfo{person}{Artem
  Rozantsev}, \bibinfo{person}{Vincent Lepetit}, {and} \bibinfo{person}{Pascal
  Fua}.} \bibinfo{year}{2016}\natexlab{}.
\newblock \showarticletitle{Direct Prediction of 3D Body Poses from Motion
  Compensated Sequences}. In \bibinfo{booktitle}{\emph{CVPR}}.
\newblock
\urldef\tempurl%
\url{https://doi.org/10.1109/CVPR.2016.113}
\showDOI{\tempurl}


\bibitem[\protect\citeauthoryear{Tome, Russell, and Agapito}{Tome
  et~al\mbox{.}}{2017}]%
        {Tome2017}
\bibfield{author}{\bibinfo{person}{Denis Tome}, \bibinfo{person}{Chris
  Russell}, {and} \bibinfo{person}{Lourdes Agapito}.}
  \bibinfo{year}{2017}\natexlab{}.
\newblock \showarticletitle{Lifting from the Deep: Convolutional 3D Pose
  Estimation from a Single Image}. In \bibinfo{booktitle}{\emph{CVPR}}.
\newblock
\urldef\tempurl%
\url{https://doi.org/10.1109/CVPR.2017.603}
\showDOI{\tempurl}


\bibitem[\protect\citeauthoryear{Vaina, Goodglass, and Daltroy}{Vaina
  et~al\mbox{.}}{1995}]%
        {Vaina1995}
\bibfield{author}{\bibinfo{person}{Lucia~M Vaina}, \bibinfo{person}{Harold
  Goodglass}, {and} \bibinfo{person}{Lawren Daltroy}.}
  \bibinfo{year}{1995}\natexlab{}.
\newblock \showarticletitle{Inference of object use from pantomimed actions by
  aphasics and patients with right hemisphere lesions}.
\newblock \bibinfo{journal}{\emph{Synthese}} \bibinfo{volume}{104},
  \bibinfo{number}{1} (\bibinfo{year}{1995}), \bibinfo{pages}{43--57}.
\newblock
\urldef\tempurl%
\url{https://doi.org/10.1007/BF01063674}
\showDOI{\tempurl}


\bibitem[\protect\citeauthoryear{Wang, Ho, Shum, and Zhu}{Wang
  et~al\mbox{.}}{2019a}]%
        {WangMonifold2019}
\bibfield{author}{\bibinfo{person}{He Wang}, \bibinfo{person}{Edmond S.~L. Ho},
  \bibinfo{person}{Hubert P.~H. Shum}, {and} \bibinfo{person}{Zhanxing Zhu}.}
  \bibinfo{year}{2019}\natexlab{a}.
\newblock \showarticletitle{Spatio-temporal Manifold Learning for Human Motions
  via Long-horizon Modeling}.
\newblock \bibinfo{journal}{\emph{IEEE Trans. Visualization \& Computer
  Graphics}} (\bibinfo{year}{2019}), \bibinfo{pages}{1--1}.
\newblock
\urldef\tempurl%
\url{https://doi.org/10.1109/TVCG.2019.2936810}
\showDOI{\tempurl}


\bibitem[\protect\citeauthoryear{Wang, Sun, Liu, Sarma, Bronstein, and
  Solomon}{Wang et~al\mbox{.}}{2019b}]%
        {Wang2019}
\bibfield{author}{\bibinfo{person}{Yue Wang}, \bibinfo{person}{Yongbin Sun},
  \bibinfo{person}{Ziwei Liu}, \bibinfo{person}{Sanjay~E. Sarma},
  \bibinfo{person}{Michael~M. Bronstein}, {and} \bibinfo{person}{Justin~M.
  Solomon}.} \bibinfo{year}{2019}\natexlab{b}.
\newblock \showarticletitle{Dynamic Graph CNN for Learning on Point Clouds}.
\newblock \bibinfo{journal}{\emph{ACM Trans. on Graphics}}
  (\bibinfo{year}{2019}).
\newblock
\urldef\tempurl%
\url{https://doi.org/10.1145/3326362}
\showDOI{\tempurl}


\bibitem[\protect\citeauthoryear{Wei, Ramakrishna, Kanade, and Sheikh}{Wei
  et~al\mbox{.}}{2016}]%
        {Wei2016}
\bibfield{author}{\bibinfo{person}{Shih-En Wei}, \bibinfo{person}{Varun
  Ramakrishna}, \bibinfo{person}{Takeo Kanade}, {and} \bibinfo{person}{Yaser
  Sheikh}.} \bibinfo{year}{2016}\natexlab{}.
\newblock \showarticletitle{Convolutional Pose Machines}. In
  \bibinfo{booktitle}{\emph{CVPR}}.
\newblock
\urldef\tempurl%
\url{https://doi.org/10.1109/CVPR.2016.511}
\showDOI{\tempurl}


\bibitem[\protect\citeauthoryear{Wu, Lim, Zhang, Tenenbaum, and Freeman}{Wu
  et~al\mbox{.}}{2016}]%
        {Wu2016}
\bibfield{author}{\bibinfo{person}{Jiajun Wu}, \bibinfo{person}{Joseph Lim},
  \bibinfo{person}{Hongyi Zhang}, \bibinfo{person}{Joshua Tenenbaum}, {and}
  \bibinfo{person}{William Freeman}.} \bibinfo{year}{2016}\natexlab{}.
\newblock \showarticletitle{Physics 101: Learning Physical Object Properties
  from Unlabeled Videos}. \bibinfo{pages}{39.1--39.12}.
\newblock
\urldef\tempurl%
\url{https://doi.org/10.5244/C.30.39}
\showDOI{\tempurl}


\bibitem[\protect\citeauthoryear{Wu, Yildirim, Lim, Freeman, and Tenenbaum}{Wu
  et~al\mbox{.}}{2015}]%
        {Wu2015}
\bibfield{author}{\bibinfo{person}{Jiajun Wu}, \bibinfo{person}{Ilker
  Yildirim}, \bibinfo{person}{Joseph~J. Lim}, \bibinfo{person}{William~T.
  Freeman}, {and} \bibinfo{person}{Joshua~B. Tenenbaum}.}
  \bibinfo{year}{2015}\natexlab{}.
\newblock \showarticletitle{Galileo: Perceiving Physical Object Properties by
  Integrating a Physics Engine with Deep Learning}. \bibinfo{pages}{127–135}.
\newblock
\urldef\tempurl%
\url{https://doi.org/10.5555/2969239.2969254}
\showDOI{\tempurl}


\bibitem[\protect\citeauthoryear{Xia, Wang, Chai, and Hodgins}{Xia
  et~al\mbox{.}}{2015}]%
        {Xia2015}
\bibfield{author}{\bibinfo{person}{Shihong Xia}, \bibinfo{person}{Congyi Wang},
  \bibinfo{person}{Jinxiang Chai}, {and} \bibinfo{person}{Jessica Hodgins}.}
  \bibinfo{year}{2015}\natexlab{}.
\newblock \showarticletitle{Realtime Style Transfer for Unlabeled Heterogeneous
  Human Motion}.
\newblock \bibinfo{journal}{\emph{ACM Trans. on Graphics (Proc. of SIGGRAPH)}}
  \bibinfo{volume}{34}, \bibinfo{number}{4}, Article \bibinfo{articleno}{119}
  (\bibinfo{date}{July} \bibinfo{year}{2015}).
\newblock
\showISSN{0730-0301}
\urldef\tempurl%
\url{https://doi.org/10.1145/2766999}
\showDOI{\tempurl}


\bibitem[\protect\citeauthoryear{Yan, Xiong, and Lin}{Yan
  et~al\mbox{.}}{2018}]%
        {Yan2018}
\bibfield{author}{\bibinfo{person}{Sijie Yan}, \bibinfo{person}{Yuanjun Xiong},
  {and} \bibinfo{person}{Dahua Lin}.} \bibinfo{year}{2018}\natexlab{}.
\newblock \showarticletitle{Spatial Temporal Graph Convolutional Networks for
  Skeleton-Based Action Recognition}. In \bibinfo{booktitle}{\emph{AAAI Conf.
  on Artificial Intelligence}}.
\newblock


\bibitem[\protect\citeauthoryear{Yao and Fei-Fei}{Yao and Fei-Fei}{2010}]%
        {Yao2010}
\bibfield{author}{\bibinfo{person}{Bangpeng Yao} {and} \bibinfo{person}{Li
  Fei-Fei}.} \bibinfo{year}{2010}\natexlab{}.
\newblock \showarticletitle{Modeling mutual context of object and human pose in
  human-object interaction activities}. In \bibinfo{booktitle}{\emph{CVPR}}.
\newblock
\urldef\tempurl%
\url{https://doi.org/10.1109/CVPR.2010.5540235}
\showDOI{\tempurl}


\bibitem[\protect\citeauthoryear{Yumer and Mitra}{Yumer and Mitra}{2016}]%
        {Yumer2016}
\bibfield{author}{\bibinfo{person}{M.~Ersin Yumer} {and}
  \bibinfo{person}{Niloy~J. Mitra}.} \bibinfo{year}{2016}\natexlab{}.
\newblock \showarticletitle{Spectral Style Transfer for Human Motion between
  Independent Actions}.
\newblock \bibinfo{journal}{\emph{ACM Trans. on Graphics (Proc. of SIGGRAPH)}}
  \bibinfo{volume}{35}, \bibinfo{number}{4}, Article \bibinfo{articleno}{137}
  (\bibinfo{date}{July} \bibinfo{year}{2016}).
\newblock
\urldef\tempurl%
\url{https://doi.org/10.1145/2897824.2925955}
\showDOI{\tempurl}


\bibitem[\protect\citeauthoryear{Zhang, Xue, Lan, Zeng, Gao, and Zheng}{Zhang
  et~al\mbox{.}}{2018}]%
        {zhang2018}
\bibfield{author}{\bibinfo{person}{Pengfei Zhang}, \bibinfo{person}{Jianru
  Xue}, \bibinfo{person}{Cuiling Lan}, \bibinfo{person}{Wenjun Zeng},
  \bibinfo{person}{Zhanning Gao}, {and} \bibinfo{person}{Narming Zheng}.}
  \bibinfo{year}{2018}\natexlab{}.
\newblock \showarticletitle{Adding Attentiveness to the Neurons in Recurrent
  Neural Networks}. In \bibinfo{booktitle}{\emph{ECCV}}.
\newblock
\urldef\tempurl%
\url{https://doi.org/10.1007/978-3-030-01240-3_9}
\showDOI{\tempurl}


\end{thebibliography}

\appendix
\section{Interaction motion dataset collection}

\noindent 
\textbf{\textsc{Walking.}} The experiment on Walking aims for estimating \emph{the width} of the path. Each subject was asked to walk back and forth on three straight paths of different widths. We simulated the width of a path using line markers to indicate path borders, and asked the subjects do not cross the borders. So we have a total of $3\times 2\times 100 = 600 $ motion samples.

\noindent 
\textbf{\textsc{Fishing.}} The experiment on Fishing aims for estimating \emph{the length} of a fishing rod. Each subject was asked to use a fishing rod to fetch a magnetic object placed in front. The object would attach to the rod's end when being touched. Each subject did 3 trails, with fishing rods of three different lengths. We have a total of $3\times  3\times 100 = 900 $ motion samples.

\noindent 
\textbf{\textsc{Pouring.}} The experiment on Pouring aims for estimating \emph{the type} of liquid. Each subject was asked to pour liquid from a cup to other one. Each subject did 3 trails with three different substances (water, shampoo, and rice). The pouring motions were effected by the viscosity or particle granularity.

\noindent 
\textbf{\textsc{Bending.}} The experiment on Bending aims for estimating \emph{the stiffness} of a power twister. Each subject was asked to bend a power twister with three different setting, from easy to hard mode.

\noindent 
\textbf{\textsc{Sitting.}} The experiment on Sitting aims for estimating \emph{the softness} of a chair being sat on. Each subject was asked to sit on four chairs of same height but different softness. The hardest chair is made of plastic, and the softest one is a yoga ball.

\noindent 
\textbf{\textsc{Drinking.}} The experiment on drinking aims for estimating \emph{the amount} of water inside a cup. Each subject was asked to take a cup from a table and get a sip of water. Each subject did 3 trails while the amount of water in the cup changed from almost full, to half full, and to almost empty. 

\noindent 
\textbf{\textsc{LiftingBox.}} The experiment on LiftingBox aims to estimate \emph{the weight} of an object from the human motion interaction. Each subject was asked to perform four different tasks in a row: (i)~lifting a box from the ground to a sofa; (ii)~lifting the box from the sofa to a table; (iii)~lifting the box from the table to the top of a closet; finally (iv)~putting the box back to the floor. Without letting the subject know,  the weight of the carrying box was randomly changed by putting different weight plates into the concealed box, ranging from 0kg to 25kg in a step of 5kg. That is, each subject  needed to do 6 trails and did not know if he/she would lift a heavy or light box before each trial, so all the captured motions are naturally close to what happens in our real life. This lifting experiment provides us $1343$ motion samples in total, all annotated with the specific task and weight. 
When a subject failed to lift up a heavy box to somewhere high, he/she did not need to perform the following tasks along the line with the same weight.

\noindent 
\textbf{\textsc{MovingBowl.}} The experiment on MovingBowl aims to judge \emph{the fragility} of an object from human motion interactions. While the weight belongs to a physical property, the fragility leans more to an empirical property. Each subject was asked to perform the similar four tasks in a row as described above, but to move a bowl this time rather than lifting a box. Three same uncovered bowls were used: one empty, one fully filled with rice, and one fully filled with water. That is, each subject was needed to do 3 trails and saw clearly the different states of these three bowls. They were all required to try their best to move the bowls without any spillage. We expect this to capture how cautious the subject was for the target task and how much that correlates to his/her motion in the corresponding trial. The degree of caution should be the highest when moving a bowl full of water, and the lowest when moving an empty bowl, which in turn relates to the level of fragility of an object. 
All action samples are annotated with one of the three levels of interacting object fragility.

\end{document}